\documentclass[10pt,journal,compsoc]{IEEEtran}
\usepackage{graphicx}
\usepackage{times}
\usepackage{array}
\usepackage{makecell}
\usepackage{graphicx}
\usepackage{tabularx}
\usepackage[hidelinks]{hyperref}
\usepackage{url}
\usepackage{multirow, multicol}
\usepackage{float}
\usepackage{booktabs}
\usepackage{tabulary}
\usepackage{longtable}
\usepackage{bm}
\usepackage{amsmath}
\usepackage{amssymb}
\usepackage{amsthm}
\usepackage{color}
\usepackage{subfigure}
\usepackage{algorithm}
\usepackage{algorithmic}
\usepackage{stmaryrd}
\usepackage{dblfloatfix}
\usepackage{diagbox}
\usepackage[american]{babel}
\usepackage{pifont}
\usepackage{tikz}
\usepackage{tikz-qtree}
\usetikzlibrary{trees}
\usetikzlibrary{positioning}
\usepackage{threeparttable}
\usepackage{enumitem}
\usepackage{csquotes}
\usepackage{colortbl}
\usepackage{caption}
\usepackage{breqn}
\usepackage{bbding}
\usepackage{footnote}

\usepackage[capitalize]{cleveref}
\crefname{section}{Sec.}{Secs.}
\Crefname{section}{Section}{Sections}
\Crefname{table}{Table}{Tables}
\crefname{table}{Tab.}{Tabs.}

\def\eqref#1{equation~\ref{#1}}

\def\1{\bm{1}}

\def\c{{\boldsymbol c}}

\def\x{{\boldsymbol x}}
\def\y{{\boldsymbol y}}
\def\z{{\boldsymbol z}}
\def\w{{\boldsymbol w}}
\def\W{{\boldsymbol W}}

\def\Y{{\boldsymbol Y}}
\def\X{{\boldsymbol X}}
\def\Z{{\boldsymbol Z}}

\definecolor{Gray}{gray}{0.9}
\definecolor{LightCyan}{rgb}{0.88,1,1}

\definecolor{lavenderblue}{rgb}{0.8, 0.8, 1}

\newtheorem{defn}{Definition}

\newcolumntype{a}{>{\columncolor{Gray}}c}
\newcolumntype{b}{>{\columncolor{white}}c}

\newcommand*{\belowrulesepcolor}[1]{%
  \noalign{%
    \kern-\belowrulesep 
    \begingroup 
      \color{#1}%
      \hrule height\belowrulesep 
    \endgroup 
  }%
} 
\newcommand*{\aboverulesepcolor}[1]{%
  \noalign{%
    \begingroup 
      \color{#1}%
      \hrule height\aboverulesep 
    \endgroup 
    \kern-\aboverulesep 
  }%
}

\ifCLASSOPTIONcompsoc
\usepackage[nocompress]{cite}
\else
\usepackage{cite}
\fi
\ifCLASSINFOpdf
\else
\fi

\begin{document}
	%
	\title{A Survey on Extreme Multi-label Learning}
	
	%
	%
	\author{Tong Wei,
		Zhen Mao,
		Jiang-Xin Shi,
		Yu-Feng Li,
		~and Min-Ling Zhang
		\IEEEcompsocitemizethanks{\IEEEcompsocthanksitem Tong Wei, Zhen Mao, and Min-Ling Zhang are with the School of Computer Science and Engineering, Southeast University, Nanjing 210096, China, and the Key Laboratory of Computer Network and Information Integration (Southeast University), Ministry of Education, China.\protect\\
			E-mail: \{weit, zhangml\}@seu.edu.cn
			\IEEEcompsocthanksitem Jiang-Xin Shi and Yu-Feng Li are with the National Key Laboratory for Novel Software Technology, Nanjing University, Nanjing, Jiangsu 210023, China.\protect\\
			E-mail: \{shijx, liyf\}@lamda.nju.edu.cn
		}
		\thanks{Manuscript received April 19, 2005; revised August 26, 2015.
	}}

%
%

\markboth{Journal of \LaTeX\ Class Files,~Vol.~14, No.~8, August~2015}%
{Wei \MakeLowercase{\textit{et al.}}: Combating Label Noise under Long-Tail Distribution}
%


\IEEEtitleabstractindextext{%
\begin{abstract}
Multi-label learning has attracted significant attention from both academic and industry field in recent decades. Although existing multi-label learning algorithms achieved good performance in various tasks, they implicitly assume the size of target label space is not huge, which can be restrictive for real-world scenarios. Moreover, it is infeasible to directly adapt them to extremely large label space because of the compute and memory overhead. Therefore, eXtreme Multi-label Learning (XML) is becoming an important task and many effective approaches are proposed. To fully understand XML, we conduct a survey study in this paper. We first clarify a formal definition for XML from the perspective of supervised learning. Then, based on different model architectures and challenges of the problem, we provide a thorough discussion of the advantages and disadvantages of each category of methods. For the benefit of conducting empirical studies, we collect abundant resources regarding XML, including code implementations, and useful tools. Lastly, we propose possible research directions in XML, such as new evaluation metrics, the tail label problem, and weakly supervised XML.
\end{abstract}

}

\maketitle

\IEEEdisplaynontitleabstractindextext

%
\IEEEpeerreviewmaketitle

\IEEEraisesectionheading{\section{Introduction}\label{sec:introduction}}
\IEEEPARstart{M}{ulti-label} learning~\cite{zhang2014review,hsu2009multi,DBLP:journals/corr/abs-1901-00248,wei2018safeml} is one of the most important machine learning paradigms, in which each real-world object is represented by a single instance (feature vector) and associated with multiple labels. Many multi-label learning algorithms have been proposed during the last decades. For example, Binary Relevance~\cite{zhang2018br} learns a binary classifier for each label separately, which ignores the label relationships. ECC~\cite{DBLP:journals/ml/ReadPHF11} learns a separate classifier for each label in a sequential manner, which means the subsequently learned classifiers can leverage previous label information. RAKEL~\cite{tsoumakas2007random} transforms the task of multi-label learning into the task of multi-class classification by mapping a random subset of labels to natural numbers, i.e., $2^{|\mathcal{Y}|} \rightarrow \mathbb{N}$, which is able to model high-order correlations between labels. With the rapid growth of training data, deep learning is vastly used to fully exploit label correlations~\cite{DBLP:conf/sigir/LiuCWY17,yuanaccelerating,xun2020correlation}.

Although classical multi-label learning is prevailing and successful, these methods typically assume that the size of label set is small, which can be too restrictive for real-world scenarios, as many applications might be complicated and have extremely large number of labels. For example, in webpage categorization~\cite{partalas2015lshtc}, millions of labels (categories) are collected in Wikipedia and one needs to annotate a new webpage with relevant labels from such a big candidate set; in image annotation~\cite{deng2009imagenet}, millions of tags are in the repository and one wishes to tag each individual picture from such a big candidate tags; in recommender systems~\cite{mcauley2015inferring}, millions of items are presented and one hopes to make informative personalized recommendations from the big candidate items. In such scenario, traditional multi-label learning methods are intractable due to the huge computational overhead. To resolve this issue, extreme multi-label learning is emerging in recent years.

\begin{figure}[!ht]
\centering
\includegraphics[width=0.495\linewidth]{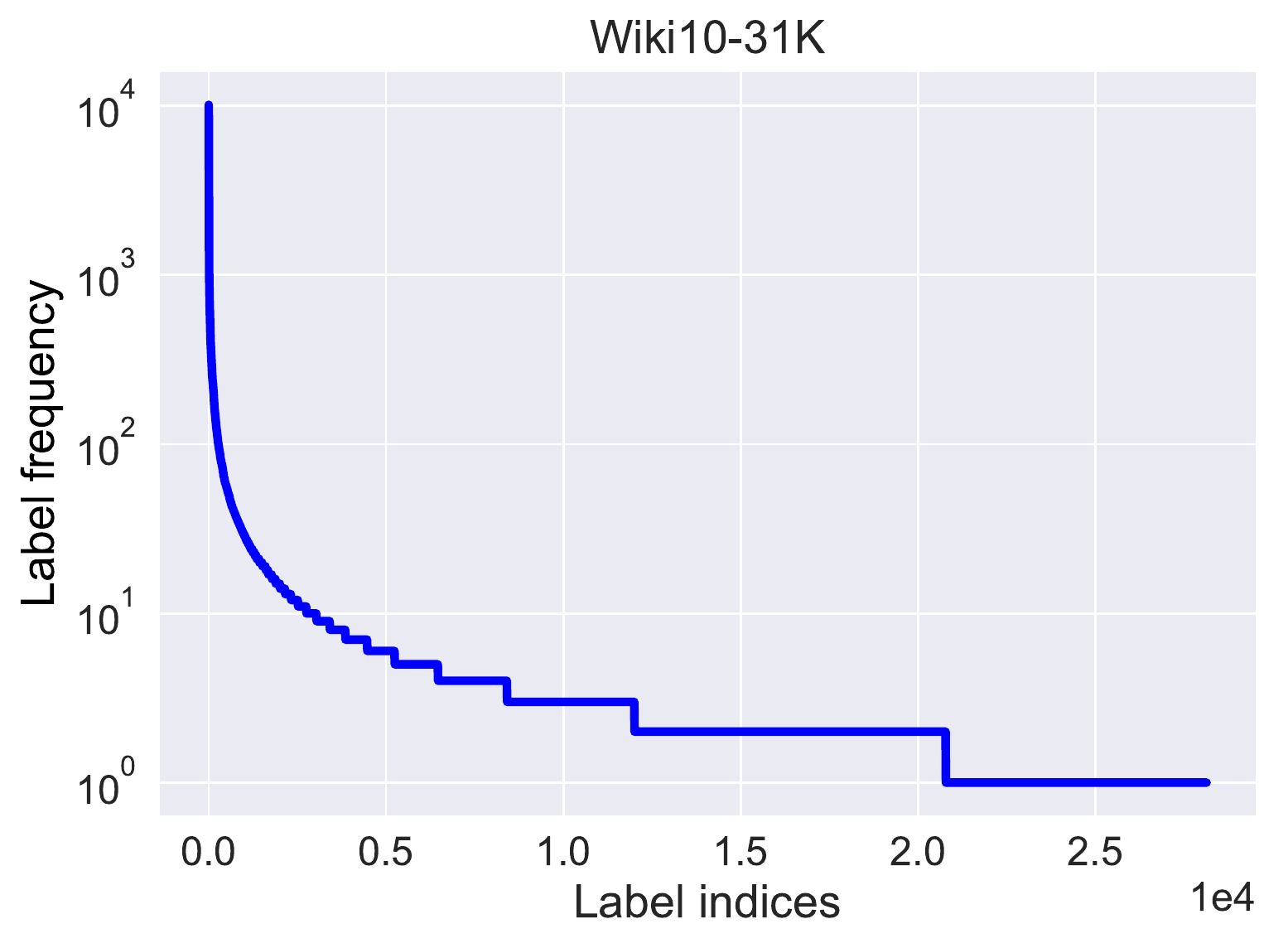}
\includegraphics[width=0.495\linewidth]{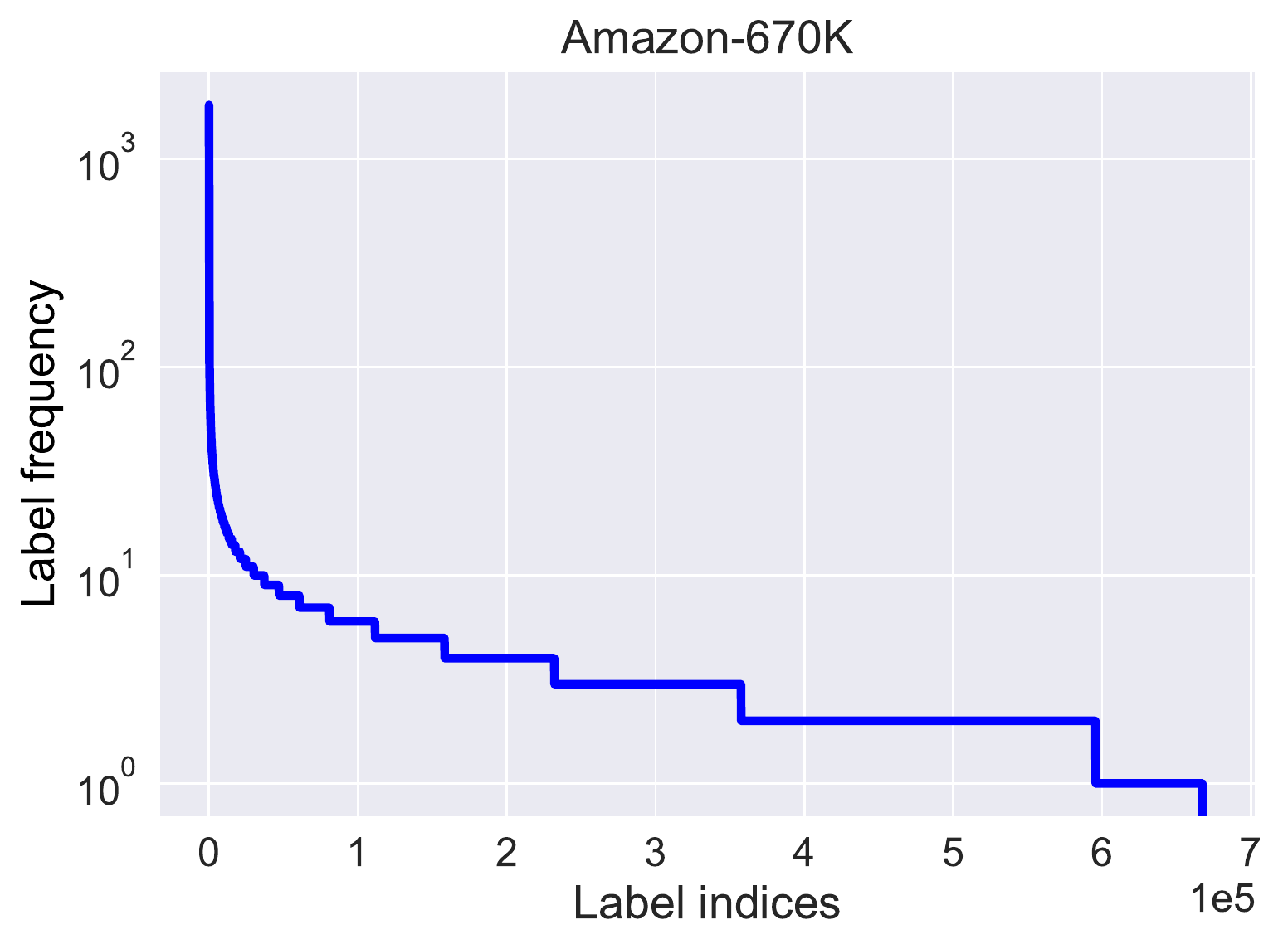}
\caption{Label frequency of Wiki10-31K and Amazon-670K datasets, which follows a long-tailed distribution.}
\label{fig:label-frequency}
\end{figure}

Extreme Multi-label Learning (XML) aims to annotate objects with the relevant labels from an extremely large number of candidate labels. XML recently owns many real-world applications~\cite{Bengio19,DBLP:journals/cacm/Varma19} such as recommender systems and search engines. Particularly, \Cref{fig:label-frequency} illustrates two real-world XML datasets of Wikipedia and Amazon which have a large number of labels whose frequencies usually follow a long-tailed distribution. Owing to the high dimensionality of label space, traditional multi-label learning approaches, such as ML-KNN~\cite{zhang2007mlknn}, RAKEL~\cite{tsoumakas2007random}, ECC~\cite{DBLP:journals/ml/ReadPHF11}, Lead~\cite{zhang2010labeld}, Binary Relevance~\cite{zhang2018br}, become infeasible and new algorithms are required. Moreover, when working with long-tailed data, this problem becomes more severe. Without taking of the long-tailed label distribution, the model performance on tail labels are far from satisfactory. Moreover, other challenges such as memory overhead and missing labels also hinders the applications of XML. Fortunately, during the past decade, XML has gradually attracted significant attentions from machine learning, data mining and related communities and has been widely applied to diverse problems~\cite{weston2011wsabie,agrawal2013multi,yu2014leml,Prabhu2014fastxml,NIPS2015sleec,babbar2016learning,yen2016pd,xu2016reml,jain2016propensity,yeh2017learning,babbar2017dismec,yen2017ppd,babbar2018adversarial,tagami2017annexml,shen2018coh,shen2018multilabel,shen2018ijcai,DBLP:journals/cacm/Varma19,wei2019ijcai,xun2020correlation}.
Specifically, in recent eight years (2014-2022), there are more than 50 papers with keyword ``extreme multi-label'' (or ``large-scale multi-label'') in the paper appearing in major machine learning and data mining conferences (including ICML/ECML-PKDD/IJCAI/AAAI/KDD/ICLR/NeurIPS). Therefore, it is important to understand existing works for both researcher and practioner, and analyze future directions in this field.


\begin{table*}[!h]
\small
\caption{Examples of applications from Amazon and Bing.}
\centering
\begin{tabular}{p{4cm}p{3cm}p{5cm}p{2cm}}
\toprule
\toprule
Source & Input sample & Output label & Output size \\
\midrule
Amazon (Prod2Query)~\cite{chang2019x} & \raisebox{-\totalheight+0.2cm}{\includegraphics[width=1.8cm, height=16mm]{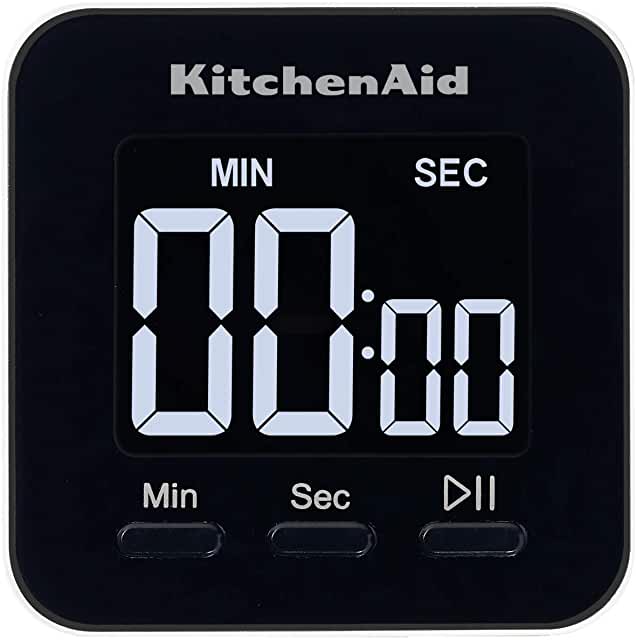}}  & 
\begin{itemize}[leftmargin=*, topsep=-1cm]
	\item black timer 
	\item black kitchen timer 
	\item kitchenaid kitchen timer
\end{itemize}
& 1 Million\\
\midrule
Bing (Related searches)~\cite{DBLP:conf/wsdm/JainBCV19} & blend gray hair &
\begin{itemize}[leftmargin=*, topsep=-1cm]
	\item highlights to blend gray hair
	\item best way to blend gray
	\item grey hair blending for women
\end{itemize}	
& 101 Million\\
\bottomrule
\bottomrule
\end{tabular}
\end{table*}

\subsection{Real-world Examples}
\subsubsection{Recommender Systems}
In Amazon product search~\cite{medini2019extreme} and Alibaba retail~\cite{song2020large}, each product is considered as a label and one may want to recommend a list of products for users which they might like from a huge candidate set. In such setting, users' information is collected as input features and their shopping history can be used to construct observed labels. It is easy to see this problem can be formulated as a multi-label learning task. As the number of products and users on platforms are huge, the training and inference speed are very demanding. In addition, personalized recommendation is desired  and the learned models should not be biased towards popular products, while seldom recommend rare products. This poses challenges to XML in recommender systems.

\subsubsection{Search Engines}
In search engines, such as Bing~\cite{DBLP:conf/wsdm/JainBCV19}, the problem of recommending related queries can be reformulated as an extreme multi-label learning task. 
After the user submits a query, the search engine need to recommend the most related queries that might serve the user's requirements from a huge candidate set. Since the size of the label set can be up to millions, existing ranking algorithms suffer from inadmissible computational cost. Therefore, designing proper ranking algorithms for search engines can be an extremely challenging task.

\subsection{Motivation and Contribution}
As far as we know, this is the first pioneer survey related to XML. The contributions of this survey can be summarized as follows:
\begin{enumerate}
\item Comprehensive review. We provide a comprehensive review of XML, including core challenges and their corresponding solutions.
\item New taxonomy. We propose a taxonomy of XML, which categorizes existing methods from three different perspectives: 1) model architecture; 2) tail-label learning; 3) weakly supervised XML.
\item Abundant resources. We collect abundant resources on XML, including open-source implementations of XML methods, datasets, tools, and paper lists.
\item Future directions. We discuss and analyze the limitations of existing XML methods. Also, we suggest possible future research directions.
\end{enumerate}

\subsection{Organization of the Survey}
The rest of the survey is organized as follows. Section~\ref{sec-overview} is an overview of XML including the background concepts, the comparison between XML and other related settings, and the core challenges of XML. Section~\ref{sec-taxonomy} introduces existing XML algorithms from three perspectives. Section~\ref{sec-experiments} lists the commonly used datasets, evaluation metrics, and resources. Section~\ref{sec-future-work} discusses the current challenges and suggests future several promising directions. Finally, Section~\ref{sec-conclude} concludes the survey.

\section{An Overview of Extreme Multi-label Learning}\label{sec-overview}

\subsection{Problem Definition of XML}
We describe notations used through the paper and the definition of XML. Let $\X = \{\x_i\}_{i=1}^N, \Y = \{\y_i\}_{i=1}^N$ be a training set of size $N$, where $\y_i$ is the label vector for data point $\x_i$. Formally, XML is the task of learning a function $f$ that maps an input (or instance) $\x \in \mathbb{R}^D$ to its target $\y \in \{0, 1\}^L$. We denote $n_j = \sum_{i=1}^N\y_{ij}$ as the frequency of the $j$-th label. Without loss of generality, we assume that the labels are sorted by cardinality in decreasing order, i.e., if $j < k$, then $n_j \geq n_k$, where $1 \leq j, k \leq L$. In our setting, we have $n_1 \gg n_L$. According to the label frequency, we can split the label set into head labels and tail labels by a threshold $\tau \in (0, 1)$. We denote head label set $\mathcal{H} = \{1, \dots, \lfloor\tau L\rfloor\}$, and tail label set $\mathcal{T} = \{\lfloor\tau L\rfloor + 1, \dots, L\}$. In practice, $\tau$ is a user-specified parameter. We summarize frequently used notations and their definitions in Table~\ref{tab:notation} for quick reference.

\subsection{Comparison with Other Settings}
To make the difference more clear, we compare XML with several existing machine learning settings in Table~\ref{tab:setting-comparison}. It is easy to see that XML differs from other settings in many aspects, causing problems in directly applying previous algorithms to XML data.

\begin{table}[h]
\centering
\caption{Commonly used notations.}
\label{tab:notation}
\begin{tabular}{ll}
\toprule
\toprule
Notations & Definition\\
\midrule
$N$ & number of training samples\\
$D$ & feature dimension\\
$L$ & size of label set\\
$\mathcal{X}$ & feature space\\
$\mathcal{Y}$ & label space\\
$\X \in \mathbb{R}^{N\times D}$ & feature matrix of training samples\\
$\Y \in \{0, 1\}^{N\times L}$ & label matrix of training samples\\
$\w \in \mathbb{R}^D$ & linear model parameters\\
\bottomrule
\bottomrule
\end{tabular}
\end{table}

\begin{table*}[!h]
\small
\caption{Comparison XML with other machine learning settings.}
\label{tab:setting-comparison}
\centering
\begin{tabular}{lllcll}
\toprule
\toprule
\backslashbox{Setting}{Characteristic}
& Feature size & Train Size & Label Size & \# Tail Label  & Label Distribution \\
\midrule
Traditional Classification~\cite{hsu2002comparison,hastie2009multi}  & Small &  Medium   & Small & Few & Balanced\\
Few-shot Learning~\cite{vinyals2016matching,snell2017prototypical} & Small& Small & Small& All & Balanced \\
Transfer Learning~\cite{pan2009survey,weiss2016survey} & Small & Small & Small& Few & Balanced\\
Imbalanced Classification~\cite{DBLP:journals/tkde/HeG09,DBLP:journals/pami/DongGZ19} & Small  & Medium  & Small  & Medium & Imbalanced\\
Traditional Multi-label learning~\cite{zhang2007mlknn,zhang2014review} & Small& Medium & Medium & Few & Balanced\\
\midrule
Extreme Multi-label Learning & Large & Large & Large &  Many & Imbalanced\\
\bottomrule
\bottomrule
\end{tabular}
\end{table*}


\subsection{The Challenges of XML}
In this section, we analyze the main challenges of XML from three perspectives, i.e., volume, quantity, and quality. The challenges involve both the input data and output labels when dealing with XML problems, which are further explained as follows.
\begin{enumerate}
\item \textit{Volume} refers to explosive growth in the amount of data, which poses many challenges to XML. First, both input feature space and output label space can grow extremely large, causing scalability problems. Second, models learned on XML data often suffers from storage overhead.
\item \textit{Quantity} refers to the frequency of labels observed in training data, which typically follows a long-tailed distribution~\cite{anderson2018longtail}. In other words, the label frequency distribution is highly imbalanced and infrequently occurred labels, which are hard to learn, account for most of the labels.
\item \textit{Quality} refers to the inferior quality of the annotated output labels. Owing to the large label set and sample size, it is very costly to annotate each instance with its complete relevant subset of labels, causing label missing problem.
\end{enumerate}

First, for the \textit{Volume} challenge, existing methods usually rely on different assumptions on the label space. The three most common assumptions are label independent assumption, low-rank assumption, and the hierarchical structure assumption. These three assumptions correspond to three types of XML algorithms, i.e., binary relevance, embedding-based methods, and tree-based method, respectively. Additionally, there are several additional techniques to further reduce the computational and memory cost, which we will cover in the next section. 

Next, for the \textit{Quantity} challenge, it means the long-tailed label distribution on both training and test datasets. Most labels occur only a few times which are referred to as \textit{tail label}, and improving the generalization on tail labels is one of the core problem for XML. To deal with tail labels, many XML methods have been proposed by various techniques, such as  the propensity-scored loss function, data enhancement, and knowledge transfer. 

Finally, for the \textit{Quality} challenge, it means that the training data suffers from weak supervision. For instance, the presence of missing labels has attracted much attention in recent years. A few works on handling missing labels use unbiased loss functions to mitigate the influence of missing labels. However, learning from weakly supervised XML data is still an open challenge which has not been well-solved.

%

\section{Taxonomy of XML}\label{sec-taxonomy}
Algorithm development always stands as the core issue of machine learning researches, with XML being no exception. During the past decade, a large number of algorithms have been proposed to learning from extreme multi-labeled data. According to the properties of each algorithm and the key challenges of XML, we propose a new taxonomy of XML, i.e., 1) model architecture, 2) tail-label learning; 3) weak supervision. Considering that it is infeasible to go through all existing algorithms within limited space, in this review we opt for scrutinizing representative XML algorithms in each line of research direction. 

\definecolor{mycolor}{RGB}{253,245,230}
\definecolor{edge}{RGB}{225,69,0}
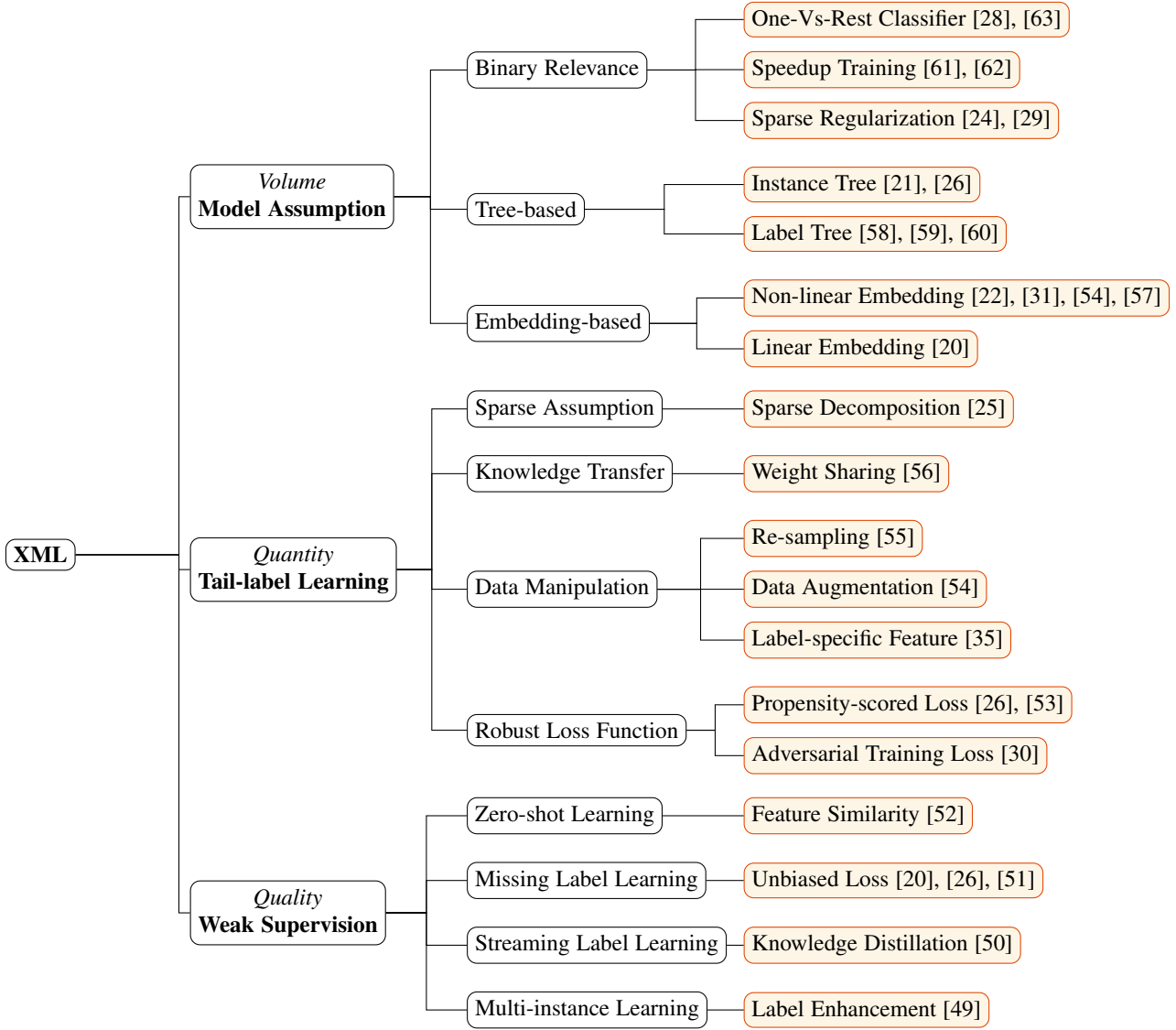
\begin{figure*}[!h]
\centering
\begin{tikzpicture}[every tree node/.style={ draw=black,rounded corners}, 
every leaf node/.style={draw=edge,fill=mycolor,rounded corners}, 
every node/.append style={align=center},
sibling distance=.2cm, grow = right, level distance = 12em, anchor = west, edge from parent fork right,
level 1/.style={sibling distance=1em, level distance = 8em},level 2/.style={sibling distance=1.2em, level distance = 12em},level 3/.style={level distance = 12em} ]

\Tree[.\textbf{XML} 
[.{\textit{Quality}\\ \textbf{Weak Supervision}}
[.{Multi-instance Learning} {Label Enhancement~\cite{shen2020mixml}}  ]
[.{Streaming Label Learning} {Knowledge Distillation~\cite{icml2020_dsll}}  ]
[.{Missing Label Learning}  {Unbiased Loss~\cite{yu2014leml,jain2016propensity,DBLP:journals/corr/abs-2007-00237}} ]
[.{Zero-shot Learning} {Feature Similarity~\cite{gupta2021generalized}}  ]
]
[.{\textit{Quantity}\\ \textbf{Tail-label Learning}}
[.{Robust Loss Function} {Adversarial Training Loss~\cite{babbar2018adversarial}}
{Propensity-scored Loss~\cite{jain2016propensity,DBLP:journals/corr/abs-1811-08812}}
]
[.{Data Manipulation}
{Label-specific Feature~\cite{wei2019ijcai}}
{Data Augmentation~\cite{chuanguo2019nips}}
{Re-sampling~\cite{DBLP:journals/corr/abs-1106-1813}}
]
[.{Knowledge Transfer} {Weight Sharing~\cite{dahiya2021deepxml}}
]
[.{Sparse Assumption} {Sparse Decomposition~\cite{xu2016reml}}
]
]
[.{ \textit{Volume}\\ \textbf{Model Assumption}}
[.{Embedding-based} {Linear Embedding~\cite{yu2014leml}} {Non-linear Embedding~\cite{NIPS2015sleec,tagami2017annexml,zhang2017deepxml,chuanguo2019nips}} ]
[.{Tree-based} 
{Label Tree~\cite{icml2018siblini,khandagale2019bonsai,you2018attentionxml}} 
{Instance Tree~\cite{Prabhu2014fastxml,jain2016propensity}}  ]
[.{Binary Relevance} 
{Sparse Regularization~\cite{yen2016pd,yen2017ppd}}
{Speedup Training~\cite{DBLP:conf/iclr/BamlerM20,DBLP:conf/aistats/ReddiKYHCK19}}
{One-Vs-Rest Classifier~\cite{babbar2017dismec,prabhu2018parabel}}
]
]
]

]
\end{tikzpicture}
\caption{Taxonomy of XML with Representative Examples.}\label{fig:model-assumption}
\end{figure*}

\begin{figure*}[!h]
\centering
\includegraphics[width=0.8\textwidth]{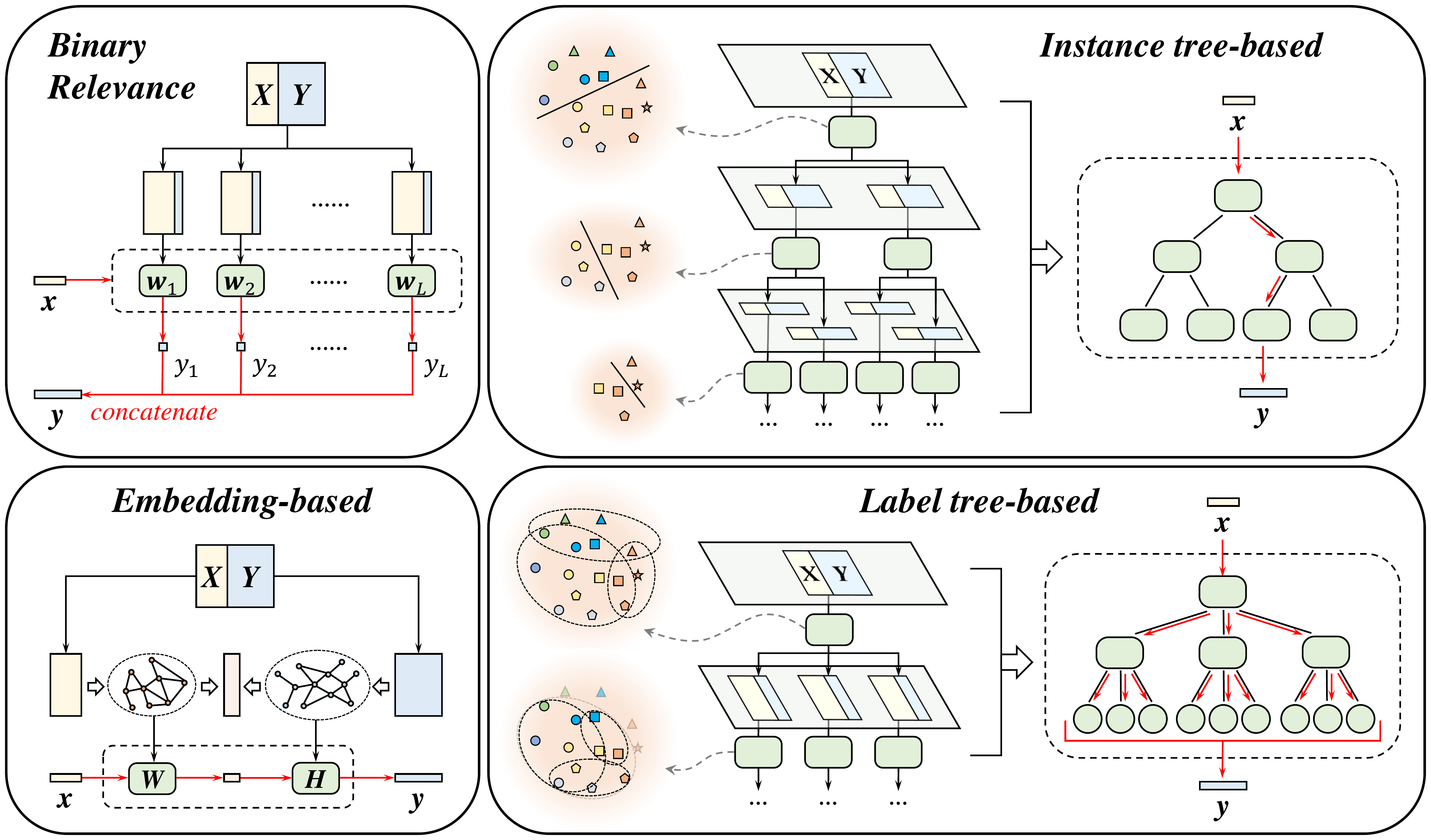}
\caption{Illustration of the core idea of different types of methods for solving XML problem, i.e., binary relevance, embedding based method, and two kinds of tree-based method.}
\label{fig:methods}
\end{figure*}

\subsection{Model Architecture}
According to the angle of solving the XML problem, most of the methods can be divided into three branches: binary relevance, embedding-based method, and tree-based method. We provide a illustration in \Cref{fig:model-assumption}.

\subsubsection{Binary Relevance}
Binary Relevance (BR) assumes that labels are independent to each other and it learns a binary classifier for each label separately~\cite{babbar2013flat,zhang2018br}. Formally, for the $j$-th label, a linear classifier is usually learned by minimizing:
\begin{equation}\label{equ:linear-classifier}
\min _{\boldsymbol{w}_{j}} \left\|\boldsymbol{w}_{j}\right\|_{2}^{2}+C \sum_{i=1}^{N} \ell \left(\boldsymbol{w}_{j}^{\top}\x_{i}, \y_{i}\right),
\end{equation}
where, $\boldsymbol{w}_j$ denotes the weight vector of the $j$-th label classifier. $\ell(\cdot, \cdot)$ is the selected loss function, such as the squared hinge loss. Notice that Problem~\eqref{equ:linear-classifier} can be efficiently solved using Liblinear~\cite{fan2008liblinear}.

We can see that the time and memory complexities of BR scale linearly in the number of labels, hence vanilla BR method suffers from high computational overhead. To this end, techniques such as parallelization~\cite{babbar2017dismec,babbar2018adversarial}, label partitioning~\cite{khandagale2019bonsai}, feature transformation~\cite{boutsidis2009columnsubsetselection,DBLP:conf/ijcai/JalanK19}, and classifier warm-up\cite{DBLP:conf/sdm/FangCHF19} are used to facilitate efficient training and testing. An interesting work named LTLS~\cite{jasinska2016log} constructs a directed acyclic graph (DAG) with $\mathcal{O}(\log_2 L)$ edges that contains exactly $L$ paths from a source vertex to a sink vertex. Every edge in the graph is associated with a learnable function. Every class corresponds to a path and the model predicts a subset of classes with the highest scoring paths. This idea can significantly reduce the training time. Further, W-LTLS~\cite{Evron2018nips} is an improved version of LTLS~\cite{jasinska2016log}, which can be seen as a special case of Error-Correcting Output Coding (ECOC)~\cite{dietterich1994ecoc}. In addition to the logarithmic inference time and model size benefiting from the trellis graph, W-LTLS has theoretical bounds developed based on previous work on ECOC. Interestingly, both LTLS~\cite{jasinska2016log} and vanilla BR can be seen as special cases of W-LTLS.

Recently, label filter~\cite{niculescu2016labelfilter} was proposed to improve the testing efficiency. When assigning labels to a test instance, most multi-label classifiers systematically evaluate every single label to decide whether it is relevant or not. Instead, this label filter pre-selects a fairly modest set of candidate labels before the base multi-label classifier is applied.


Additionally, to alleviate memory overhead, recent works restrict the model capacity by imposing sparse constraints~\cite{yen2016pd,yen2017ppd}, removing spurious parameters~\cite{babbar2017dismec,babbar2018adversarial,khandagale2019bonsai}, and Count-min Sketch technique~\cite{medini2019extreme}. In particular, the $\ell_2$ regularizer in Problem~(\ref{equ:linear-classifier}) is substituted by $\ell_1$ regularizer, i.e., $||\boldsymbol{w}||_1$. Therefore, PD-Sparse~\cite{yen2016pd} shows that a margin-maximizing loss with $\ell_1$ penalty, in case of extreme classification, yields extremely sparse solution both in primal and in dual without sacrificing the expressive power of predictor. The objective is as follows.
\begin{equation}
\min_{\boldsymbol{W}}  \sum_{j=1}^{L}\left\|\boldsymbol{w}_{j}\right\|_{1}+ C \sum_{i=1}^{N} \ell\left(\boldsymbol{w}^{\top} \boldsymbol{x}_{i}, \boldsymbol{y}_{i}\right)
\end{equation}

\underline{\textbf{Remark:}}
Though the rationale behind BR is simple, it often achieves performance very close to sophisticated XML approaches, including deep learning methods, which hence raises an interesting question ``does label correlation matter in XML?''. Even for extant deep learning methods, very few approaches explicitly leverage label correlations. Using techniques such as GNNs~\cite{kipf2016semi} to model the label correlation may be a promising direction and can help answer the above question. Additionally, the idea of BR classifiers are used in many tree-based approaches.

\begin{table}[!h]
\caption{A summary of training, testing, and memory complexities of techniques which are incorporated with Binary Relevance to deal with XML challenges. We respectively denote $N, D, L$ as size of training set, feature set, and label set. $\boldsymbol{W}$ denotes the model weight matrix. $k$ is the parameter for parallelization.}
\label{tab:br-methods-summary}
\centering
\begin{tabular}{l|ccc}
\toprule
\toprule
\backslashbox{Technique}{Metric}
& \makecell{Train} & \makecell{Test} & \makecell{Memory}   \\
\midrule
Vanilla BR & $\mathcal{O}(NDL)$ & $\mathcal{O}(DL)$ & $\mathcal{O}(DL)$  \\
Label Filter~\cite{niculescu2016labelfilter} & $\mathcal{O}(NDL)$ & $\mathcal{O}(D\hat{L})$ & $\mathcal{O}(DL)$ \\
Label Pruning~\cite{DBLP:conf/aaai/WeiL19} & $\mathcal{O}(ND\hat{L})$  & $\mathcal{O}(D\hat{L})$ &  $\mathcal{O}(D\hat{L})$ \\
Weight Pruning~\cite{babbar2017dismec,DBLP:conf/aaai/WeiL19} & $\mathcal{O}(NDL)$ & $\mathcal{O}(||\boldsymbol{W}||_0)$ & $\mathcal{O}(||\boldsymbol{W}||_0)$ \\
Parallelization~\cite{babbar2017dismec} & $\mathcal{O}(ND\frac{L}{k})$  & $\mathcal{O}(D\frac{L}{k})$ &$\mathcal{O}(DL)$ 	 \\
Negative Sampling~\cite{DBLP:conf/wsdm/JainBCV19} & $\mathcal{O}(\hat{N}DL)$ & $\mathcal{O}(DL)$& $\mathcal{O}(DL)$  \\
Dense Feature~\cite{DBLP:conf/wsdm/JainBCV19} & $\mathcal{O}(N\hat{D}L)$ & $\mathcal{O}( \hat{D}L)$ & $\mathcal{O}(\hat{D}L)$ \\
\bottomrule
\bottomrule
\end{tabular}
\end{table}

\subsubsection{Embedding-based methods} 
Embedding-based methods usually assume that the label matrix is low-rank which can exploit the underlying correlations between labels. There are typically two types of projection methods, i.e., the linear or non-linear projection, which seek to embed the feature space and label space onto a joint low-dimensional space. Then, a model is learned in this low-dimensional hidden space~\cite{tai2012multilabel,chen2012feature,yu2014leml,NIPS2015sleec,tagami2017annexml,Evron2018nips,chuanguo2019nips}. This line of methods can dramatically reduce the model parameters compared with the plain BR methods, but usually involve solving complex optimization problems.

For example, LEML~\cite{yu2014leml} is one of the earliest works hat propose to explore the low-dimensional projection. It attempts to make training and prediction tractable by assuming that the training label matrix is low-rank and reducing the effective number of labels by projecting the high dimensional label vectors onto a low dimensional linear subspace. The objective of LEML is formulated as:
\begin{equation}
\begin{aligned}
\min _\Z \sum_{i=1}^N \sum_{j=1}^L \ell\left(y_{i j}, f^j\left(\boldsymbol{x}_i ; \Z\right)\right)+\lambda \cdot r(\Z)
\text { s.t. } \operatorname{rank}(\Z) \leq k \text {, }
\end{aligned}
\end{equation}
where $\Z \in \mathbb{R}^{D \times L}$ is the parameters of the projection matrix, $f(\x, \Z) = \Z^\top \boldsymbol{x}$, and $r(\Z)$ is the trace norm regularizer. Consider a low-rank decomposition of the form $\Z=\W \boldsymbol{H}^\top$, where $\W \in \mathbb{R}^{D \times k}$ and $\boldsymbol{H} \in \mathbb{R}^{L \times k}$, this optimization problem can be solved by applying alternating minimization for $\boldsymbol{W}$ and $\boldsymbol{H}$. By learning the low-rank projection, LEML increases scalability and efficiency.

However, linear embedding methods typical suffer from limited expressive capability, SLEEC~\cite{NIPS2015sleec} learns local distance preserving embeddings which can accurately predict infrequently occurring (tail) labels. This allows SLEEC to break free of the traditional low-rank assumption and boost classification accuracy by learning embeddings which preserve pairwise distances between only the nearest label vectors. The objective of SLEEC is formulated as:
\begin{equation}
\min _{\Z \in \mathbb{R}^{\tilde{L} \times N}}\left\|P_{\Omega}\left(\Y^{\top} \Y\right)-P_{\Omega}\left(\Z^{\top} \Z\right)\right\|_{F}^{2}+\lambda\|\Z\|_{1}.
\end{equation}
Motivated by SLEEC, AnnexML~\cite{tagami2017annexml} presents a novel graph embedding method which constructs a $k$-NN graph of label vectors and attempts to reproduce the graph structure in the embedding space. The prediction is efficiently performed by using an approximate nearest neighbor search (ANNS)~\cite{dong2011efficient} that efficiently explores the learned k-nearest neighbor graph in the embedding space. Lately, SML~\cite{DBLP:conf/icml/LiuS19} proves theoretical properties of SLEEC. 

Since it is natural to use deep neuron networks to learn embeddings for its powerful expressive capability, C2AE~\cite{yeh2017learning} performs joint feature and label embedding by deriving a deep latent space, followed by the introduction of label-correlation sensitive loss function for recovering the predicted label outputs. Similarly, Rank-AE~\cite{DBLP:conf/naacl/WangCSQLZ19} proposes the ranking-based AutoEncoder~\cite{vincent2010stacked} to simultaneously explore inter-label dependencies and the feature-label dependencies by projecting labels and features onto a common embedding space. DeepXML~\cite{zhang2017deepxml} explores the label space by building and modeling an explicit label graph and learn non-linear embedding for both features and labels. An obvious difference between XML and other text classification problems is that labels have the same form of text as instances. SiameseXML~\cite{dahiya2021siamesexml} takes full advantage of that. A siamese network is leveraged to encode both instance and label features. The inner product of instance and label features can be used as a probability estimate for the instance that belongs to the label. As part of the classifier, the label feature alleviates the problem of insufficient number of tail label samples and reduces the number of parameters and training costs of the model.

Recently, GNNs~\cite{kipf2016semi} have shown promising performance at capturing label correlations and GNN-XML~\cite{zong2020gnn}, ECLARE~\cite{mittal2021eclare} and GalaXC~\cite{saini2021galaxc} all use GNNs to model label correlations for XML. GNNs augment features by aggregating the representations of neighbor nodes. GalaXC~\cite{saini2021galaxc} utilizes the label-document graph to augment document feature through GNN and attention mechanism. This alleviates the long tail problem in XML. ECLARE~\cite{mittal2021eclare} builds a dynamic label graph using random walks. This graph is leveraged to augment label feature and assist shortlisting.  BGNN-XML~\cite{zong2022bgnn} proposes
a bilateral-branch graph isomorphism network to decouple representation learning and classifier learning for better modeling tail labels and use an adaptive learning strategy to avoid heavy over-fitting on tail labels and insufficient training on head labels.

While embedding-based methods usually enjoy theoretical properties, its generalization performance is typically outperformed by improved BR and tree-based methods. GLaS~\cite{chuanguo2019nips} empirically and theoretically demonstrates that embedding-based methods often severely overfit training data, which results in poor generalization performance. To this end, they propose a regularization term combining the spread-out and Laplacian regularizers. 
\begin{multline}
\sum_{y \in L_{\boldsymbol{y}}} \sum_{y^{\prime} \notin L_{\boldsymbol{y}}}\left[f(\x)_{y^{\prime}}-f(\x)_{y}+c\right]_{+} +\\ \frac{1}{K^{2}}\left\|\boldsymbol{V}^{\top} \boldsymbol{V}-\frac{1}{2}\left(\boldsymbol{A} \Z^{-1}+\Z^{-1} \boldsymbol{A}\right)\right\|_{F}^{2},
\end{multline}
where $\boldsymbol{A}=\Y^{\top} \Y$, $\Z = \operatorname{diag}(\boldsymbol{A}) \in \mathbb{R}^{K \times K}$ and $\boldsymbol{V}$ is the label embedding matrix.
Maximum inner product search~\cite{DBLP:journals/corr/AuvolatV15} is performed in inference. The performance w.r.t. PSP@$k$ over tail labels is dramatically improved.

Unlike tree-based methods, embedding-based methods usually require negative sampling to improve training efficiency and achieve positive/negative examples balance. However, memory overheads of popular negative mining techniques often force mini-batch sizes to remain small and slow training down. To alleviate this problem, methods like SiameseXML~\cite{dahiya2021siamesexml}, GalaXC~\cite{saini2021galaxc} obtain their negative examples from both mini-batch and the global memory buffer. NGAME~\cite{DBLP:journals/corr/abs-2207-04452} combines min-batch and hard negative sampling and uses a Negative Mining-aware Mini-batching technique, which allows much bigger batch size.  

\underline{\textbf{Remark:}}
To sum up, conventional embedding-based methods have many advantages including simplicity, ease of implementation, strong theoretical foundations, the ability to handle label correlations, the ability to adapt to online and incremental scenarios, etc. Unfortunately, embedding methods can also pay a heavy price in terms of prediction accuracy due to the loss of information during the compression phase. Notably, most deep learning methods can also be viewed as embedding-based methods which learns latent representations by neural networks which is more expressive than linear projection used in conventional embedding-based methods. Although the theoretical foundations of deep learning need further study, it can be employed to improve classification performance.

\subsubsection{Tree-based methods}
In comparison to other types of approaches, tree-based methods greatly reduce inference time, which generally scales logarithmically in the number of labels. There are typically two types of trees including instance trees~\cite{Prabhu2014fastxml,jain2016propensity,icml2018siblini,DBLP:conf/aistats/MajzoubiC20} and label trees~\cite{weston2013labelpartition,weston2013labelpartition,daume2016recalltree,Liang2018block,you2018attentionxml,DBLP:conf/sigir/WydmuchJBD21,yu2022pecos,zhang2021fast}, depending whether instance or label is partitioned in tree nodes. We respectively introduce these two types tree-based methods in the following.

\textbf{Instance Tree}
In instance trees, each node consists of a set of training examples and then distributed to the child nodes. The intuition is that each region of feature space contains only a small number of active labels. The active label set at a node is the union of the labels of all training points present in that node. As one of the most earliest works, MLRF~\cite{agrawal2013multi} developes multi-label random forests to tackle problems with millions of labels by partitioning a parent's feature space between its children so that each individual child has to deal with not only fewer training data points but also labels. The tree is grown until the number of active labels in each leaf node is logarithmic in the total number of labels. Instead of splitting nodes by calculating Gini index, FastXML~\cite{Prabhu2014fastxml} incorporates the rank-based loss to yield better performance. Specifically, it formulates a novel node partitioning objective which directly optimizes an nDCG based ranking loss and which implicitly learns balanced partitions. The objective of the split rule in FastXML is formulated as:
\begin{equation}
\begin{aligned}
\min & \|\boldsymbol{w}\|_{1}+\sum_{i} C_{\delta}\left(\delta_{i}\right) \log \left(1+e^{-\delta_{i} \boldsymbol{w}^{\top} \x_{i}}\right) \\ &-C_{r} \sum_{i} \frac{1}{2}\left(1+\delta_{i}\right) \mathcal{L}_{\mathrm{nDCG} @ k}\left(\boldsymbol{r}^{+}, \boldsymbol{y}_{i}\right) \\ &-C_{r} \sum_{i} \frac{1}{2}\left(1-\delta_{i}\right) \mathcal{L}_{\mathrm{nDCG} @ k}\left(\boldsymbol{r}^{-}, \boldsymbol{y}_{i}\right) \\ \text { w.r.t. } & \boldsymbol{w} \in \mathbb{R}^{D}, \boldsymbol{\delta} \in\{-1,+1\}^{L}, \boldsymbol{r}^{+}, \boldsymbol{r}^{-} \in \Pi(1, L).
\end{aligned}
\end{equation}

RecallTree~\cite{daume2016recalltree} creates a new online reduction of multi-label classification to binary classification for which training and prediction time scale logarithmically with the number of classes. It uses an BR-like structure to make a final prediction, but instead of scoring every class, it only scores a small subset of $\mathcal{O}(\log K)$ classes by dynamically building tree to efficiently whittle down the set of candidate classes. 
It is known that GBDT achieves more promising results in many tasks, however, it was shown that vanilla GBDT~\cite{friedman2001gbdt} can easily run out of memory or encounter near-forever running time in the XML setting, and propose a new GBDT variant,  GBDT-\textsc{sparse}\cite{si2017gbdt}, to resolve this problem by employing $\ell_{0}$ regularization. GBDT-\textsc{sparse} makes the crucial observation that each data point has very few labels; based on that it solves a $\ell_{0}$ regularized optimization problem to enforce the prediction of each leaf node in each tree to have only a small number $k$ of nonzero elements or labels. Hence, after $T$ trees have been added during GBDT iterations, there will be at most $Tk$ nonzero gradients for any data point. After that, CraftXML~\cite{icml2018siblini} proposes a new variant of random forest method. (i) It exploits a random forest~\cite{liaw2002randomforest} strategy which not only randomly reduces both the feature and the label spaces to obtain diversity but also replaces random selections with random projections to preserve more information; (ii) it uses a novel low-complexity splitting strategy which avoids the resolution of a multi-objective optimization problem at each node.

\textbf{Label Tree}
Label tree methods usually assume that labels are organized in a tree structure.
In label trees, each node consists of a set of labels which are then distributed to the child nodes. The tree structure is determined by recursively clustering labels until terminal conditions are reached. Specifically, the aim of label clustering is to split $\mathcal{Y}$ into disjoint subsets by taking label similarity into account. This is achieved by  $K$-means algorithm~\cite{lloyd1982least}. Let $\c_k, \forall k \in [K]$ denotes the set of labels of the $k$-th cluster, the objective function of label clustering can be formulated as
\begin{equation}\label{equ:kmeans}
\min _{\boldsymbol{c}_{1}, \ldots, \boldsymbol{c}_{K}} \sum_{k=1}^{K} \sum_{i \in \boldsymbol{c}_{k}} dist\left(\boldsymbol{v}_{i}, \boldsymbol{\mu}_{k}\right) 
\end{equation}
where $dist(\cdot,\cdot)$ represents a distance function and $\boldsymbol{\mu}_{k}$ denotes the center of the $k$-th cluster. The distance function is defined in terms of the cosine similarity as $dist \left(\boldsymbol{v}_{i}, \boldsymbol{\mu}_{k}\right) = 1 - \frac{\boldsymbol{v}_i^\top \boldsymbol{\mu}_k}{||\boldsymbol{v}_i|| \cdot ||\boldsymbol{\mu}_k||}$. Problem~(\ref{equ:kmeans}) is NP-hard and an approximate solution can be found using the standard $K$-means algorithm.

At test time, given a data point $\x$ and a label $y_j$ that is relevant to $\x$, we denote $e$ as the leaf node $y_j$ belongs to and $\mathcal{A}(e)$ as the set of ancestor nodes of $e$ and $e$ itself. Note that $|\mathcal{A}(e)|$ is path length from root to $e$. Denote the parent of $n$ as $p(n)$. We define the binary indicator variable $z_n$ to take value $1$ if node $n$ is visited during prediction and $0$ otherwise. From the chain rule, the probability that $y_j$ is predicted as relevant for $\x$ is as follows:
\begin{equation}
\begin{aligned}
\operatorname{Pr}\left(y_{j}=1 \mid \x\right) &=\operatorname{Pr}\left(y_{j}=1 \mid z_{e}=1, \x\right) \\
& \times \prod_{n \in \mathcal{A}(e)} \operatorname{Pr}\left(z_{n}=1 \mid z_{p(n)}=1, \x\right)
\end{aligned}
\end{equation}

In particular, HOMER~\cite{tsoumakas2008effective} propose a hierarchical multi-label classifier which recursively divides label set evenly using balanced K-means algorithm. At each node, a multi-label classifier is trained to associate examples with the successive nodes. Note that, the number of child nodes is far less than the size of label set. Therefore, this approach can be seen as a divide-and-conquer strategy. Label trees were also applied to deal with multi-class classification problems \cite{DBLP:conf/nips/BengioWG10,DBLP:conf/nips/DengSBL11} where the tree structure and node classifiers are learned from data. Further, the probabilistic label tree (PLT)  are proposed as a pruning strategy for inference when optimizing F-measure. 

Lately, AttentionXML~\cite{you2018attentionxml} incorporates Glove embedding~\cite{pennington2014glove} and the attention mechanism~\cite{DBLP:conf/emnlp/LuongPM15} into XML to learn semantic information from raw text data using LSTM~\cite{lstm1997}. To speedup the training, it builds multiple probabilistic label trees. For each tree, it trains a single conventional multi-label classifier at each level of the tree using BiLSTM~\cite{huang2015bilstm}. With respect to the depth of trees, Bonsai~\cite{khandagale2019bonsai} claims that shallow trees are superior to deeper trees.

PECOS~\cite{yu2022pecos} is a versatile and modular machine learning framework used in tree-based methods. It draws on the idea of information retrieval system and has the following three steps; (i) indexing: partitioning labels into several groups; (ii) matching: shortlisting labels that may associate with the document; (iii) ranking: scoring the matched documents. XR-Linear~\cite{yu2022pecos} and XR-Transformer~\cite{zhang2021fast} both use this framework. They recursively execute the above three stages, and the original task is divided into subtask trees, so that these subtasks can be executed efficiently. XR-Linear is a recursive linear realization, using a OVR linear classifier as its matcher in step (ii) and ranker in step (iii). XR-Transformer is a recursive realization of PECOS with transformer as its matcher. Compared with its non-recursive version X-Transformer~\cite{chang2020taming}, XR-Transformer is way more efficient to train.

A recent work~\cite{DBLP:conf/nips/LiuCYHD21} pointed out that the current tree-based methods are based on the unimodal assumption that each label only represents one specific semantic. So these methods partition each label into specific cluster. But this assumption is often invalid in practice, because labels may have multiple semantics. They proposed to disentangle the different semantics of a label by partitioning the labels into multiple clusters.

\begin{figure}
\centering
\includegraphics[width=0.5\textwidth]{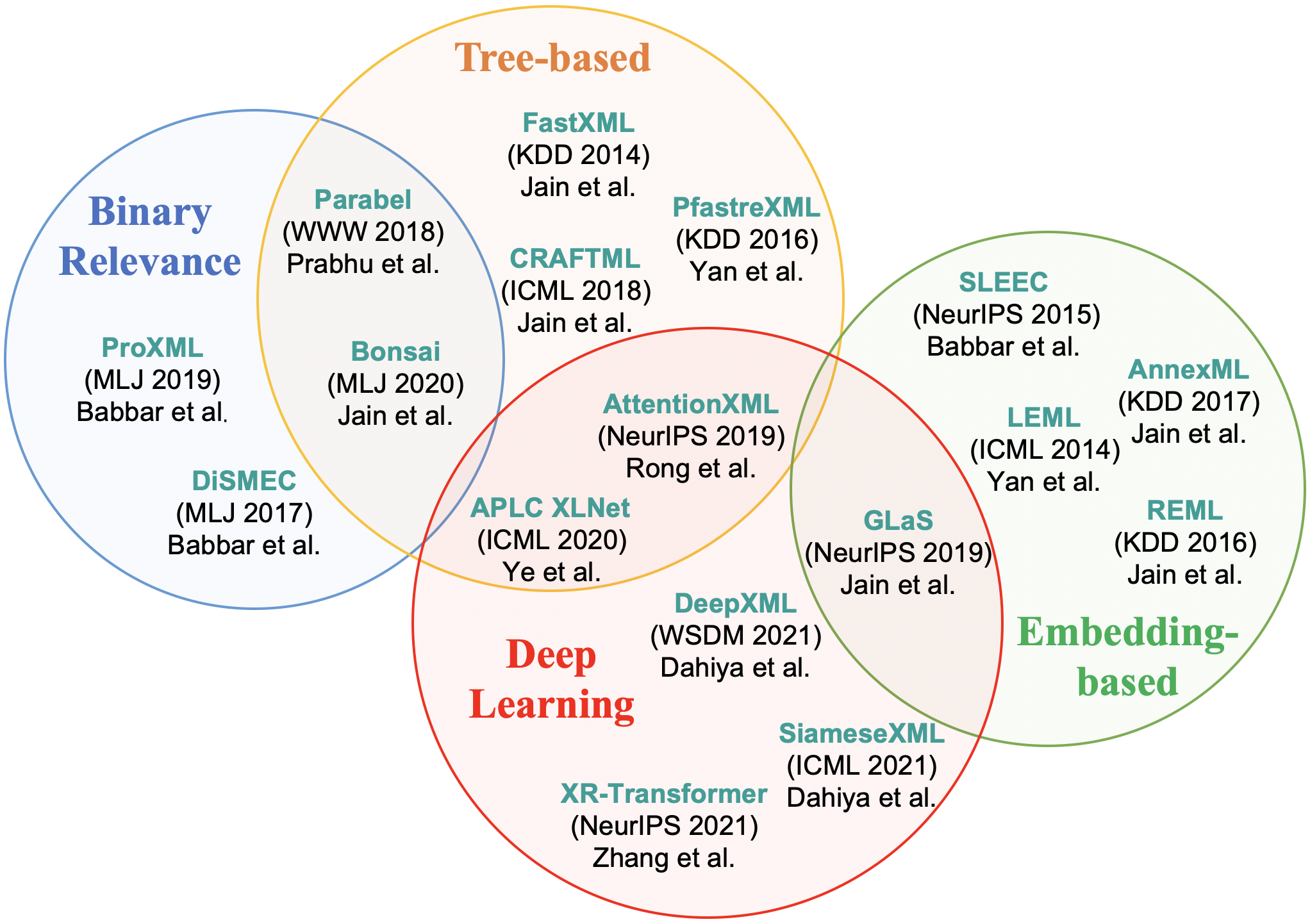}
\caption{Domain relevance of different types of XML methods.}\label{fig:domain-relevance}
\end{figure}

\underline{\textbf{Remark:}}
It is a common belief that tree-based methods usually suffer from low prediction accuracy affected by the so-called cascading effect, where the prediction error at the top cannot be corrected at a lower level. Therefore, to alleviate this issue, addition modules can be designed as a post-processing step. Hence, incorrect predictions made by tree models can be corrected by optimizing proper objectives. From the aspect of efficiency, tree methods enjoy fast prediction speed which generally scales logarithmic in the number of labels, in comparison with other types of approaches. We summarize the difference of instance trees and label trees in Table~\ref{tab:instance-label-tree-summary}. Note that the introduced XML methodologies may overlap with each other. For instance, the label trees are usually applied in deep learning methods for efficient training. An example of the relationship between different methodologies is illustrated in Figure~\ref{fig:domain-relevance}.

\begin{table}[!htbp]
\caption{A comparison of \emph{Instance Tree} and \emph{Label Tree}.}
\label{tab:instance-label-tree-summary}
\centering
\begin{tabular}{lcccc}
\toprule
\toprule
& \makecell{Width} & \makecell{Depth} & \makecell{Node classifier} & \makecell{Inference}\\
\midrule
Instance tree & Binary &  Deep  &Discriminative & Single path \\
Label tree & K-ary & Shallow  & Probabilistic & Multiple paths  \\
\bottomrule
\bottomrule
\end{tabular}
\end{table}

\subsubsection{Additional Model Speedup Techniques}
We have noticed that the efficiency of BR can be reduced by many techniques, such as imposing sparse reguarlizers and leveraging parallel computing. Additionally, there are other general methods to further reduce the training consumption, especially for deep learning approaches. We introduce the shortlisting and negative sampling in below.

\textbf{Shortlisting}
When using a document for training and inferring, the cost of estimating its association with millions of labels is especially high in XML. Also, it leads to a serious imbalance of positive and negative examples for a single label in the training phase. The shortlisting technique can resolve these problems very well. During training, it usually selects a small number (usually set to $\mathcal{O}(\log L)$) of negative labels and calculate the objective function on positive and selected negative labels, which can achieve sub-linear time consumption. Rather than randomly selecting negative label subsets, a more principle way is choosing negative labels that are possibly confusing for the model, a.k.a., hard negative labels. At test time, a subset with size $\mathcal{O}(\log L)$ of possibly related labels to the example will be selected and the classifier evaluates the scores for only the shortlisted labels to get the final prediction.

To select hard negative labels and the most relevant labels for the test point, \textit{Approximate nearest neighbor search (ANNS)}~\cite{malkov2018efficient} is a commonly used approach. It builds a similarity graph of labels using  embeddings of label descriptions. When querying for a test point, it calculates the similarity of document and label embeddings and returns the most similar subset of labels very efficiently. The shortlisting has been used in many recent deep learning approaches, such as Astec~\cite{dahiya2021deepxml} and SiameseXML~\cite{dahiya2021siamesexml}, to accelerate training and testing.

Interestingly, the label tree~\cite{choromanska2015log,prabhu2018parabel,you2018attentionxml,yu2022pecos,zhang2021fast} can be seen as a special type of shortlisting. It recursively passes the test point  from the root to the leaf. The root note is a collection of all labels and internal and leaf nodes correspond to a subset of possibly relevant labels, which has the same effect with shortlisting. Some deep learning approaches do use label trees for shortlisting and then evaluate the classifier only for shortlisted labels, such as XR-Transformer~\cite{zhang2021fast} and XR-linear~\cite{yu2022pecos}.

Additionally, DECAF~\cite{mittal2021decaf} and ECLARE~\cite{mittal2021eclare} employ a cluster-based shortlisting approach which has a balanced clustering of labels and learns the one-vs-all classifiers for each cluster. Given a test data point, labels in the clusters with the highest score are shortlisted. 

\textbf{Negative Sampling}
Similar to shortlisting, negative sampling also aims to reduce the training cost in XML problems. It chooses a small fraction of negative labels for each example when calculating objective functions. A direct sampling approach is to randomly draw negative labels from a uniform distribution over the class labels, i.e., stochastic negative sampling~\cite{DBLP:conf/aistats/ReddiKYHCK19}. Since there is no sum over all labels, it only requires $\mathcal{O}(K)$ time to calculate losses as well as gradients for each data point, where $K$ is the number of selected negative labels. 

Although uniform negative sampling is simple and easy to implement, it was demonstrated that uniform negative sampling is suboptimal because of the poor gradient signal caused by the fact that negative samples are too easy to distinguish from positive samples. Typically, a data set with many labels is comprised of several hierarchical clusters, with large clusters of generic concepts and small sub-clusters of specialized concepts. When drawing negative samples uniformly, the correct label will likely belong to a different generic concept than the negative label. The model quickly learns to assign very low scores to negative labels, making their contribution to the gradient exponentially small. To improve the uniform negative sampling, hard negative sampling was proposed to mine negative labels that are hard to distinguish from positive labels~\cite{DBLP:conf/iclr/BamlerM20}. The hard negative labels can be generated by an auxiliary model which is implemented as a balanced hierarchical binary decision tree such that drawing negative label scales only as $\mathcal{O}(\log |\mathcal{Y}|)$.

\underline{\textbf{Remark:}} 
Shortlisting and negative sampling are becoming a very commonly used technique in XML to effectively reduce the training and testing complexities. To leverage shortlisting and negative sampling, there is a trade-off between generalization performance and computational cost which might be needed to tune. There are also some room for improving existing techniques. For instance, common shortlisting approaches are based on the measure of similarities among label features to select hard negatives. However, on one hand, label similarities may not indicate the difficulty of distinguishing positive and negative labels under long-tailed distribution; on the other hand, since label similarities are calculated beforehand, it is interesting to leverage the learning dynamics of DNNs, e.g., the training sample loss or memorization effect, to mine hard negatives.

\subsection{The Long-Tailed Label Distribution}

\begin{figure}[tbp]
\centering
\begin{minipage}{.25\textwidth}
\centering
\includegraphics[width=1\textwidth]{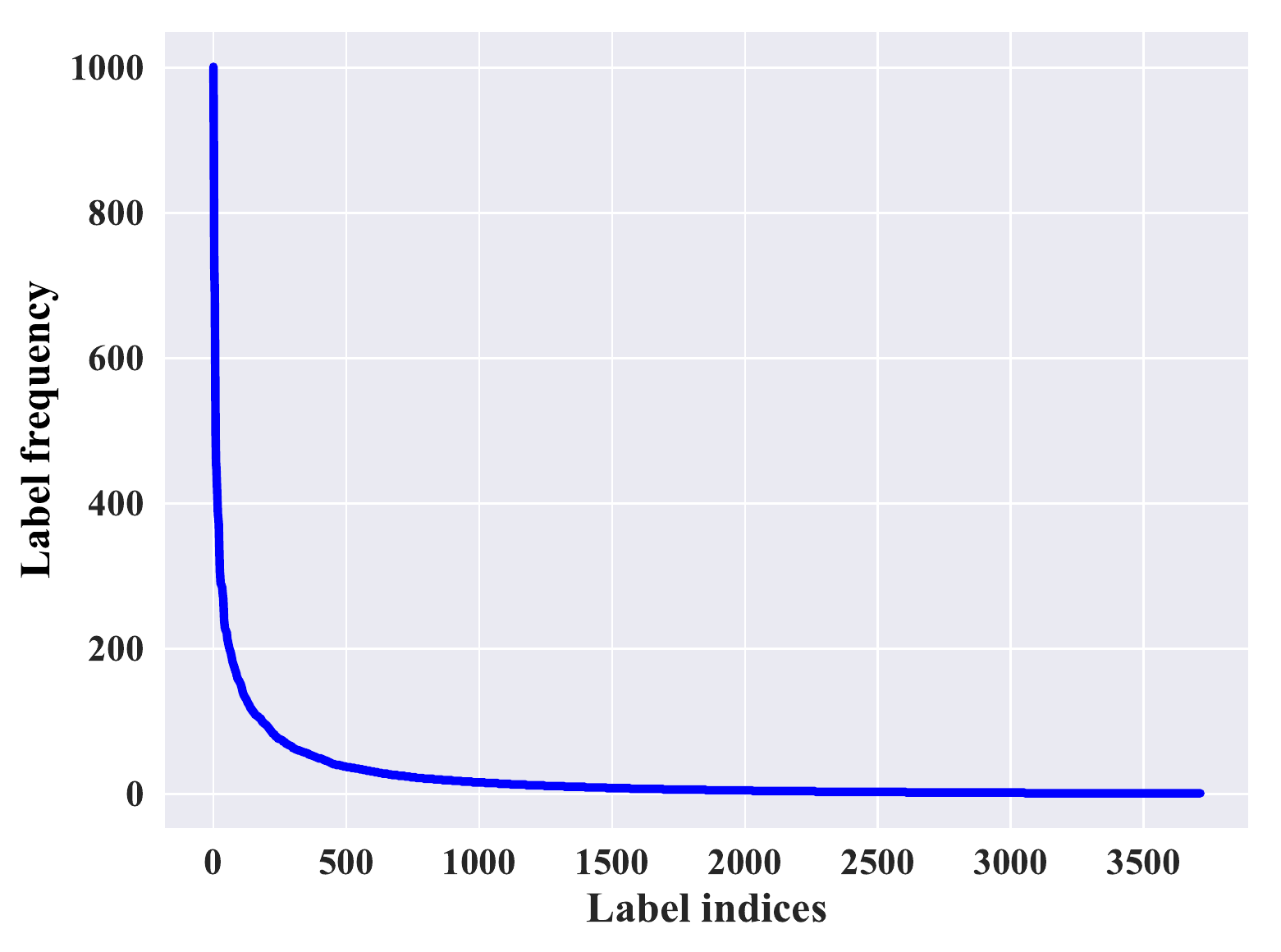}
\end{minipage}%
\begin{minipage}{.25\textwidth}
\centering
\includegraphics[width=1\textwidth]{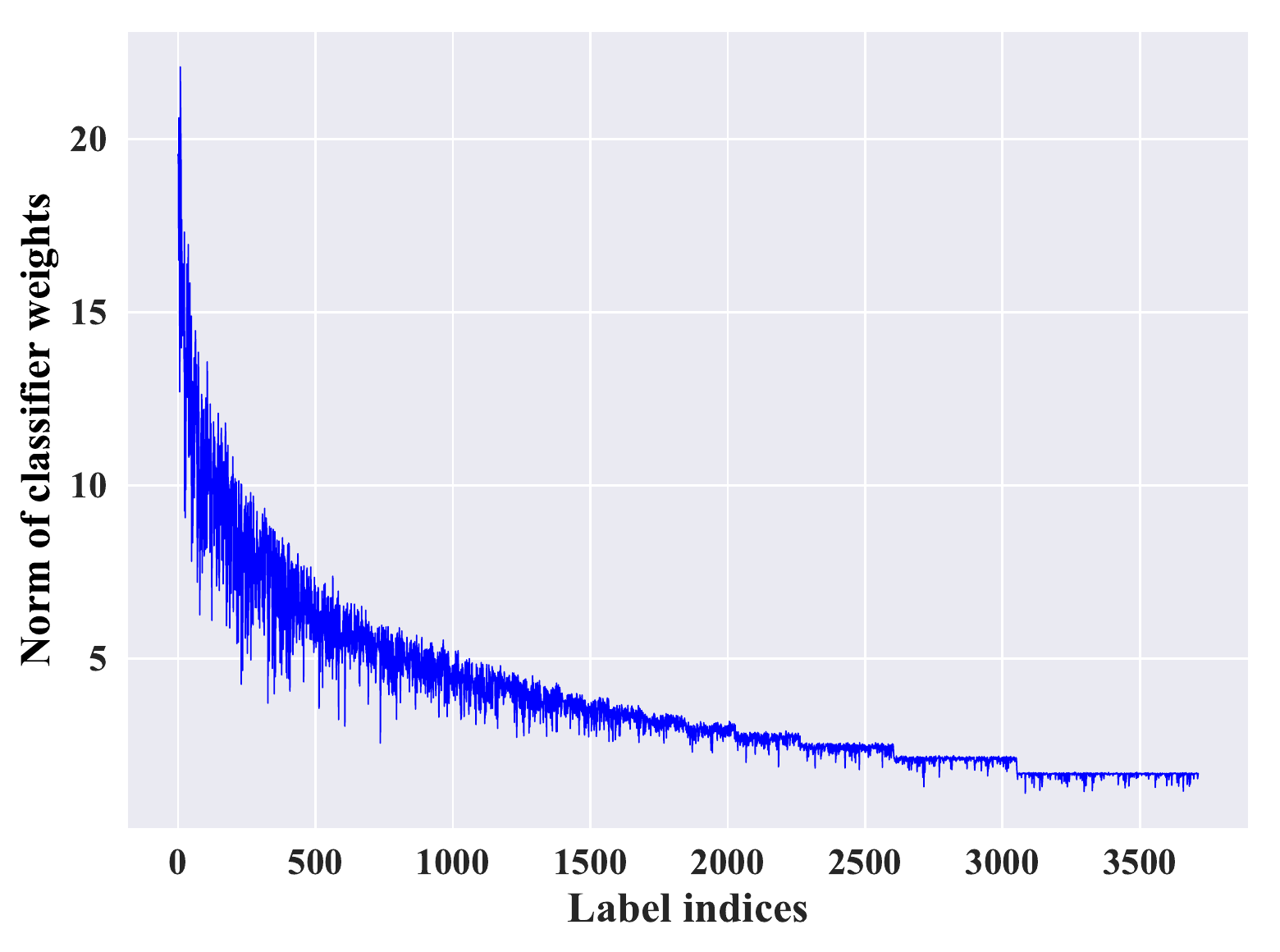}
\end{minipage}%
\caption{Left: Label frequency follows a long-tailed distribution. Right: Distribution of the norm of label classifier weights.}\label{fig:long-tail}
\end{figure}

The quotes said by Chris Anderson in his book~\cite{anderson2018longtail} is as follows.
\begin{displayquote}

The three main observations: (1) the tail of available variety is far longer
than we realize; (2) it's now within reach economically; (3) all those
niches, when aggregated, can make up a significant market.
\end{displayquote}

In XML, one important statistical characteristic is that labels follow a long-tailed distribution as illustrated in Figure~\ref{fig:long-tail}. Infrequently occurring labels (referred to as \emph{tail labels}) possess limited training examples and are harder to predict than frequently occurring ones (referred to as \emph{head labels}), with definitions given in Definition~\ref{def-head-tail}. Tail labels have recently attracted increasing attention~\cite{DBLP:journals/tip/MaoTG13,NIPS2015sleec,xu2016reml,jain2016propensity,liu2017jmlr1,babbar2018adversarial,wei2018ijcai} since tail tail label are more informative and rewarding in many real-world applications, such as personalized recommendation. Conventional approaches take all labels with equal importance and make the learned model perform better on head labels, while results in worse performance for tail labels. 

In the following, we first present a informal definition of head labels and tail labels. Then, we introduce several effective approaches to deal with tail labels.

\begin{defn} [\textbf{Head Label} \& \textbf{Tail Label}]\label{def-head-tail}
Let $\mathcal { D } = \left\{ \boldsymbol { x } _ { i } , \boldsymbol { y } _ { i } \right\} _ { i = 1 } ^ { N }$ be a large-scale multi-label dataset where labels follow a power-law distribution. Suppose labels $\{l_1, \cdots, l_K\}$ are organized by frequencies in descending order where $\sum_{ i = 1 } ^ { N }  Y_ {i,k} >= \sum_{ i = 1 } ^ { N } Y_ {i,k+1} $, $\forall 1 \leq k \leq K - 1$. By setting a threshold $h$, Frequently occurring labels $\{l_1, \cdots, l _ { h }\}$ are referred to as head labels and infrequently ones $\{l _ {h + 1 }, \cdots, l _ K\}$ are referred to as tail labels.
\end{defn}

To improve the performance of tail labels, many approaches have been studied. In the following subsections, we introduce each type of methods with its representative work.

\subsubsection{Robust Loss Function}
PfastreXML~\cite{jain2016propensity} is one of the earliest works that propose to improve tail label performance. It optimizes a propensity-scored nDCG@$k$ (a.k.a. PSnDCG@$k$) which is formalized as follows.
\begin{equation}
\begin{aligned} 
\min_{\boldsymbol{W}} & \|\boldsymbol{w}\|_{1}+C_{\delta} \sum_{i} \log \left(1+e^{-\delta_{i} \boldsymbol{w}^{\top} \x_{i}}\right) \\ &+C_{r} \sum_{i} \frac{1}{2}\left(1+\delta_{i}\right) \ell_{\mathrm{PSnDCG} @ k}\left(\boldsymbol{r}^{+}, \boldsymbol{y}_{i}\right) \\ &+C_{r} \sum_{i} \frac{1}{2}\left(1-\delta_{i}\right) \ell_{\mathrm{PSnDCG} @ k}\left(\boldsymbol{r}^{-}, \boldsymbol{y}_{i}\right) \\ \mathrm{w} . \text { r. t. } & \boldsymbol{w} \in \mathcal{R}^{D}, \boldsymbol{\delta} \in\{-1,+1\}^{L}, \boldsymbol{r}^{+}, \boldsymbol{r}^{-} \in \Pi(1, L)
\end{aligned}
\end{equation}
Here, the PSnDCG@$k$ is defined as:
\begin{equation}
	\text{PSnDCG}@k := \frac{{\text{PSDCG}}@k}{\sum_{l=1}^{k} \frac{1}{\log(l+1) } }
\end{equation}
where
$\text{PSDCG}@k := \sum_{l\in {\text{rank}}_k (\hat{\boldsymbol y})} \frac{\boldsymbol y_l}{p_l\log(l+1)}$. Notice that $p_l$ is the inverse propensity score for the $l$-th label:
$$
p_{l} = 1+C\left(n_{l}+B\right)^{-A},
$$
and $A, B, C$ are hyperparameters which can be different for different datasets.
By optimizing PSnDCG@$k$, the learned model puts higher priorities on correctly predicting tail labels than head labels, hence improving the generalization performance of tail labels.

Another interesting work REML~\cite{xu2016reml} treats tail labels as outliers and decomposes the label matrix into a low-rank matrix which depicts label correlations and a sparse one capturing the influence of tail labels. The objective of REML is formulated as:
\begin{equation}
\begin{aligned}
\min _{\hat{\Y}_{L}, \widehat{\Y}_{S}} & \left\|\Y-\widehat{\Y}_{L}-\widehat{\Y}_{S}\right\|_{F}^{2}. \\
\text { s.t. }  & \operatorname{rank}\left(\widehat{\Y}_{L}\right) \leq k, \quad \operatorname{card}\left(\widehat{\Y}_{S}\right) \leq s
\end{aligned}
\end{equation}
REML is an improved version of LEML~\cite{yu2014leml} which does not take tail label into account. This low-rank and sparse decomposition was also used in recommender systems to recommend long-tailed products~\cite{li2017coldstartandlongtail}, but  needs to solve non-trivial optimization problems which are computationally expensive.

Later, ProXML~\cite{babbar2018adversarial} regards this phenomenon as a setup in which an adversary~\cite{shafahi2019adversarial} is generating test examples such that the features of the test set instances are quite different from those in the training set. For each label $j$, its robust optimization objective function is formulated as:
\begin{equation}
\min _{\boldsymbol{w}} \max _{\left(\tilde{\x}_{1}, \ldots, \tilde{\x}_{N}\right)} \sum_{i=1}^{N} \max \left[1-s_{i}\left(\left\langle\boldsymbol{w}, \x_{i}-\tilde{\x}_{i}\right\rangle\right), 0\right],
\end{equation}
where $s_i = 2 Y_{i,j} - 1 \in \{-1, +1\}$ and $\tilde{\x}_i$ is a perturbation for instance $\x_i$.
Opposed to PfastreXML~\cite{jain2016propensity}, it claims that hamming loss is a more suitable loss function to optimize in XML scenario since it treats tail labels equally with head labels. Interestingly, the idea of ProXML is equivalent to confine the norm of classifier weights $\w$ to a small limit.

\subsubsection{Data Manipulation}
A simple idea to improve the tail label performance is to generate more data. For instance, GLaS~\cite{chuanguo2019nips} trains DNNs with input dropout which is a simple data augmentation technique for text classification models with sparse features. It  uniformly removes features in the input with probability $\rho$, which discourages the model from fitting spurious patterns in input features when training data is scarce. Formally, for a selected keep probability $\rho \in[0,1]$ and an input feature $\x$, the method produces an augmented input $\x^{\prime}=\x \odot$ Bernoulli $(\rho, D)$, where $\odot$ denotes element-wise multiplication. Thus, non-zero feature coordinates are set to zero with probability $1-\rho$. This can promote the model to be robust to corruption of the input features.

Recently, Re-Rank~\cite{DBLP:conf/kdd/WeiTLY21} finds that models trained on long-tailed datasets tend to have imbalanced label classifier norms as illustrated in~\ref{fig:long-tail}. In other words, classifiers of head labels usually have larger norms, while norms tail label classifiers are smaller. This may make a lot impact in inference because the prediction score is calculated as the dot product of label classifier and testing sample features, leading to larger prediction score for head labels as per empirical studies. To this end, they propose to balance the classifier weights for labels to improve the tail label performance through data augmentation~\cite{DBLP:conf/emnlp/WeiZ19}.

A different work LIFT~\cite{wei2019ijcai} constructs label-specific features that have special design for tail labels to improve the performance. By generating low-dimensional features, it can also improve the efficiency of the approach. A similar idea was explored DEFRAG~\cite{jalan2019defrag}.

\subsubsection{Knowledge Transfer}
Regarding transferring knowledge from some labels to some others, ECC~\cite{DBLP:journals/ml/ReadPHF11} is a simple approach. ECC trains a binary classifiers for each label sequentially. Given each label, it trains the classifier using both original features and the predictions of preceding labels' classifiers. This can effectively leverage label correlations. Following this idea, CCMC~\cite{liu2017jmlr1} develops a classifier chain model by automatically identify easy and hard classes and proposes easy-to-hard learning paradigms. Tail labels are supposed to be identified as hard labels because of the scarcity of positive training examples. Recently, DeepXML~\cite{dahiya2021deepxml} learns two deep models are trained on head labels and tail labels. The semantic representation from head labels are transferred to tail label model, which can reduce the number of examples for learning representations.

\underline{\textbf{Remark:}} 
Although the tail label problem is usually ignored in previous XML literature, more and more recent studies focus on this problem for its practicability. However, there are still big room for improvement. For instance, most existing studies are based on traditional machine learning models, well-designed deep learning approaches should achieve superior performance. Moreover, although data augmentation for tail labels are demonstrated to be effective \cite{DBLP:conf/kdd/WeiTLY21}, however, simply applying existing class-agnostic augmentation techniques~\cite{DBLP:conf/emnlp/WeiZ19} is unfavorable because they do not consider label correlations in XML and may further increase the computational cost. How to better conduct data augmentation for XML
is still an open question. Additionally, in computer vision tasks, many training strategies are proposed to deal with the issue of long-tailed label distribution including class-balanced loss function and decoupled training, it would be interesting to adapt them to XML.

\subsection{Weakly Supervised XML}
In real-world XML applications, it is costly to obtain strong supervision by annotating all data. Instead, we usually confront with a weakly supervised learning problem. For instance, due the large number of candidate labels, some relevant labels are missing. Even worse, examples for some labels may not present in the training dataset. Also, new labels may emerge over time rather than available beforehand. In some applications, we have to deal with multi-instance XML problem.

\textbf{Missing labels} As an intuitive approach, LEML~\cite{yu2014leml} handles missing labels by training model on observed labels only which means the position of missing entries in label matrix need be known in advance. Such formulation has elegant theoretical analysis, however, can not capture tail label practically. 
\begin{equation}
\begin{aligned} \min _{\Z} & \sum_{(i, j) \in \Omega} \ell \left(Y_{i j}, f^{j}\left(\boldsymbol{x}_{i} ; \Z\right)\right)+\lambda \cdot r(\Z), \\
	\text {s.t. } &\operatorname{rank}(\Z)  \leq k \end{aligned}
\end{equation}

PfastreXML~\cite{jain2016propensity} does not erroneously treat missing labels as irrelevant but
instead provide unbiased estimates of the true loss function even when ground truth labels go missing under arbitrary probabilistic label noise models. This paper addresses this issue by developing propensity scored variants of P@$k$ and nDCG@$k$ which provide unbiased estimates of the true loss as if computed on the complete ground truth without any missing labels. SwiftXML~\cite{prabhu2018extreme} also uses propensity scored losses to deal with missing labels. Later,
PW-DiSMEC~\cite{DBLP:journals/corr/abs-2007-00237} extends the loss functions in PfastreXML to anylabel decomposable loss functions~\cite{menon2019multilabel} such as squared loss, (squared) hinge loss, and binary cross-entropy.


Kanehira et al.~\cite{kanehira2016true} considered that the existence of false-negative examples can severely degrade the performance when using AUC as the optimization objective. The authors train an uni-class model and approximate false-negativeness of each examples for each label. Then use false-negativeness as another penalty term in the objective. This work presented a possible method to deal with false-negative examples.
In recommender systems, since only positive feedback are observed, Yang et al.~\cite{Yang2018recsys} proved the evaluation metric and the recommendation algorithms are biased toward popular items. The authors then propose an unbiased estimator using inverse propensity score.

An extreme case of missing labels is zero-shot learning which recently attracts a lot of attention. For instance, ZestXML~\cite{gupta2021generalized} discusses a task called Generalized Zero-shot eXtreme Multi-label Learning (GZXML), in which  the label set contains many-shot, few-shot and zero-shot labels. To deal with these unseen label, ZestXML learns a sparse weight matrix $\boldsymbol W$ and computes the relevance score $\x_i^\top W z_l$ between $i$-th document feature $\x_i$ and $l$-th label feature $\z_l$. The weight matrix $\boldsymbol W$ can be learned by the object function below:
\begin{equation}
	\begin{aligned}
		\min_{\boldsymbol{W}} &\frac12 ||{\boldsymbol{W}}||_F^2+\lambda\sum_{i=1}^{N}\sum_{l=1}^{L}\log\left(1+e^{-\y_{il}\x_i^T{\boldsymbol{W}}\z_l}\right),\\
		\text{s.t.} & ||{\boldsymbol{W}}_{i*}||_0\leq K\; \forall i \in \left\{1,...,C\right\}
	\end{aligned}
\end{equation}
where $\y_{il}$ denotes the $l$-th label of $i$-th training document, $k$ denotes the maximum number of non-zero elements in each row of $\boldsymbol{W}$. Although ZestXML can handle the GZXML problem, the training of ZestXML still needs to rely on a large amount of labeled training data.
Inspired by ZestXML~\cite{gupta2021generalized}, MACLR~\cite{xiong2021extreme} divides current XML setup into four categories, i.e., eXtreme Multi-label Learning(XML), Generalized Zero-shot Extreme Multi-label Learning (GZXML), Few-Shot eXtreme Multi-label Learning(FS-XML), Extreme Zero-shot eXtreme Multi-label Learning(EZ-XML). In EZ-XML, relations between instances and  labels are unknown, only raw text of instances and labels are accessible. FS-XML is a relaxed form of EZ-XML, where relations between instances and labels are partially known. Based on contrastive learning, MACLR~\cite{xiong2021extreme} tackles EZ-XML through a two-stage pre-training procedure. In stage $1$, the encoder is pretrained by Inverse Cloze Task~\cite{lee2019latent} with both instances and labels text. In stage $2$, the pseudo-labels are constructed according to the correlation between instance features and label features which are obtained by the encoder in stage $1$ and TF-IDF. These pseudo-labels are leveraged to further train the encoder.
More recently, GROOV~\cite{simig2022open} introduces a new scenario of XML, i.e., open vocabulary XML (OXML). In OXML, given an instance, some corresponding labels remain agnostic during training and testing. Hence, methods in this task need to generate these labels. GROOV employs a seq2seq model to generate labeled word sequences.

\underline{\textbf{Remark:}} 
As missing labels is prevalent in real-world applications, the performance of existing approaches still needs improvements. Most existing works rely on the random missing assumption which cannot approximate the realistic problem. Further studies on non-random missing are desired. Additionally, the zero-shot XML is a special case of XML with missing labels. However to simultaneous handle long-tailed label distribution and missing labels is very challenging.


\section{ Run Experiments in XML }\label{sec-experiments}

\subsection{Implementations}
To facilitate the selection of appropriate XML methods in different applications, Table~\ref{tab:compare} provides a comparison of representative methods from five perspective: P1 efficiency, P2 tail label, P3 model size, P4 missing label, P5 performance, and implementation of each method. 
For P1-P4, ``$\bigcirc$''denotes it is completely supported, ``$\triangle$ " denotes it is supported but not completely, ``\ding{53}'' denotes it is not supported. For P5, the number of ``$\star$'' indicates the generalization performance of the method, with a maximum of ``$\star$$\star$$\star$$\star$$\star$'' and a minimum of ``$\star$''.

\subsubsection{Dataset Statistics} 
The detail statistics of commonly used XML datasets are listed in Table~\ref{dataset}. All datasets are available at the XML repository~\cite{xmlrepo}. 
To read raw text data into sparse matrix format, online tools are available~\footnote{\url{https://github.com/kunaldahiya/pyxclib}}.

\begin{table*}[!h]
\caption{Data sets statistics}\label{dataset}
\centering
\begin{tabular}{lrrrrrr}
	\toprule
	\toprule
	\makecell[l]{Data set} & \makecell[c]{Train \\$N$} & \makecell[c]{Features \\$D$} & \makecell[c]{Labels \\$L$} & \makecell[c]{Test \\$M$} & \makecell[c]{Avg. labels\\ per point} & \makecell[c]{Avg. points\\ per label}\\\midrule
	Bibtex & 4,880 & 1,836 &  159 & 2,515 & 2.40 & 111.71	\\
	Delicious & 12,920 & 500 & 983 & 3,185 & 19.03 & 311.61\\
	EUR-Lex & 15,539 & 5,000 & 3,993 & 3,809 & 5.31 & 25.73\\
	Wiki10 & 14,146 & 101,938 & 30,938 & 6,616 & 18.64 & 8.52\\      
	DeliciousLarge & 196,606 & 782,585 & 205,443 & 100,095 & 75.54 & 72.29\\
	WikiLSHTC-325K & 1,778,351& 1,617,899& 325,056&587084 & 17.4& 3.2 \\
	Wiki-500K & 1,813,391 &2,381,304 &501,070 &783743&24.7& 4.7 \\
	Amazon-670K & 490,499& 135,909& 670,091& 153025 & 3.9& 5.4 \\
	\bottomrule
	\bottomrule
\end{tabular}
\end{table*}

\begin{savenotes}
\begin{table*}
	\caption{Comparison of representative XML methods with respect to five properties.}\label{tab:compare}
	\centering
		\begin{tabular}{l|l|c|c|c|c|c|c}
			\toprule
			\toprule
			\textbf{Category} & \textbf{Method} & \makecell{\textbf{P1} \\{\tiny efficiency}} & \makecell{\textbf{P2} \\ {\tiny tail label}} & \makecell{\textbf{P3}\\ {\tiny model size}} & \makecell{\textbf{P4}\\{\tiny missing label}} &  \makecell{\textbf{P5}\\{\tiny performance}} & \textbf{Implementation}\\
			\midrule
			
			\multirow{5}{*}{Embedding-based}	
			& AnnexML~\cite{tagami2017annexml} 						&    \cellcolor{lavenderblue} $\bigcirc$   &    \cellcolor{pink} \ding{53} &      \cellcolor{lavenderblue} $\bigcirc$  &   \cellcolor{pink} \ding{53}   &  $\star \star$ & Official (C++)\footnote{\url{https://github.com/yahoojapan/AnnexML}} \\
			\cline{2-2} \cline{8-8}   
			& ExMLDS~\cite{Gupta2019exmlds} 						&      \cellcolor{lavenderblue} $\bigcirc$ &    \cellcolor{pink} \ding{53}  &      \cellcolor[gray]{0.9} $\triangle$ &      \cellcolor{lavenderblue} $\bigcirc$ &     $\star\star$  & Official (C++)\footnote{\url{https://bitbucket.org/vgupta123/exmlds/src/master/}}  \\
			\cline{2-2} \cline{8-8}
			& LEML~\cite{yu2014leml} 							  &       \cellcolor{lavenderblue} $\bigcirc$ &   \cellcolor{pink} \ding{53}    &        \cellcolor[gray]{0.9} $\triangle$   &  \cellcolor{lavenderblue} $\bigcirc$ & $\star$ & Official (Python)\footnote{\url{https://github.com/AnthonyMRios/leml}}\\
			\cline{2-2} \cline{8-8}
			
			& DEFRAG~\cite{jalan2019defrag} 						&          \cellcolor{lavenderblue} $\bigcirc$ &   \cellcolor{pink} \ding{53}    &      \cellcolor{lavenderblue} $\bigcirc$ &      \cellcolor{pink} \ding{53}  &  $\star\star$ & Official (C++)\footnote{\url{https://github.com/purushottamkar/defrag}}\\
			\cline{2-2}  \cline{8-8}
			& W-LTLS~\cite{Evron2018nips}  							&          \cellcolor{lavenderblue} $\bigcirc$ &   \cellcolor{pink} \ding{53}    &      \cellcolor{lavenderblue} $\bigcirc$ &      \cellcolor{pink} \ding{53} &  $\star\star$ & Official (Python)\footnote{\url{https://github.com/ievron/wltls}}\\
			\cline{2-2}  \cline{8-8}
			& SLEEC~\cite{NIPS2015sleec} &  \cellcolor{lavenderblue} $\bigcirc$ &   \cellcolor{pink} \ding{53}    &      \cellcolor{pink} \ding{53} &      \cellcolor{pink} \ding{53}  &  $\star\star\star$ & Official (Matlab)\footnote{\url{http://manikvarma.org/code/SLEEC/download.html}}\\
			\midrule
			\multirow{5}{*}{Tree-based} 
			& LdSM~\cite{DBLP:conf/aistats/MajzoubiC20} 			&      \cellcolor{lavenderblue} $\bigcirc$ &     \cellcolor{pink} \ding{53}  &      \cellcolor[gray]{0.9} $\triangle$ &         \cellcolor{pink} \ding{53} &  $\star\star$ & Official (C++)\footnote{\url{https://github.com/mmajzoubi/LdSM}}\\
			\cline{2-2} \cline{8-8}         & PfastreXML~\cite{jain2016propensity} 					&    \cellcolor{lavenderblue} $\bigcirc$ &  \cellcolor{lavenderblue} $\bigcirc$  &    \cellcolor{lavenderblue} $\bigcirc$     &    \cellcolor[gray]{0.9} $\triangle$    & $\star\star\star$ & Official (C++,Matlab)\footnote{\url{http://manikvarma.org/code/PfastreXML/download.html}} \\
			\cline{2-2}   \cline{8-8}       & PLT~\cite{jasinska2016extreme} 			&		\cellcolor{lavenderblue} $\bigcirc$	&    \cellcolor{pink} \ding{53}   &    \cellcolor{pink} \ding{53}   &      \cellcolor{lavenderblue} $\bigcirc$         & $\star\star\star$ & Official (C++)\footnote{\url{https://github.com/mwydmuch/extweme_wabbit}}\\
			\cline{2-2}  \cline{8-8}        & FastXML~\cite{Prabhu2014fastxml} 						&  \cellcolor{lavenderblue} $\bigcirc$     &    \cellcolor{pink} \ding{53}   &     \cellcolor{lavenderblue} $\bigcirc$     &    \cellcolor{pink} \ding{53}   &  $\star\star$ & Official (Python)\footnote{\url{https://github.com/Refefer/fastxml}}\\
			\cline{2-2}  \cline{8-8}        & SwiftXML~\cite{prabhu2018extreme}  					&       \cellcolor{lavenderblue} $\bigcirc$ &  \cellcolor{lavenderblue} $\bigcirc$ &             \cellcolor[gray]{0.9} $\triangle$ &       \cellcolor[gray]{0.9} $\triangle$ & $\star\star\star$ & Official (C++,Matlab)\footnote{\url{http://manikvarma.org/code/SwiftXML/download.html}}\\
			\cline{2-2}  \cline{8-8} & CRAFTML~\cite{icml2018siblini} & \cellcolor{lavenderblue} $\bigcirc$  & \cellcolor{pink} \ding{53} & \cellcolor{lavenderblue} $\bigcirc$ & \cellcolor{pink} \ding{53} & $\star\star\star$ & Official (Java)\footnote{\url{https://github.com/Orange-OpenSource/OpenCraftML}}\\
			\midrule
			\multirow{7}{*}{Binary Relevance} 
			& DiSMEC~\cite{babbar2017dismec} 						&      \cellcolor{lavenderblue} $\bigcirc$ &      \cellcolor{pink} \ding{53} &     \cellcolor{lavenderblue} $\bigcirc$  &         \cellcolor{pink} \ding{53}  &  $\star\star\star\star$ & Official (C++)\footnote{\url{https://github.com/xmc-aalto/dismec}}\\
			\cline{2-2} \cline{8-8}         & Label Filters~\cite{niculescu2016labelfilter} 		&      \cellcolor{lavenderblue} $\bigcirc$ &    \cellcolor{pink} \ding{53}   &      \cellcolor{pink} \ding{53} &          \cellcolor{pink} \ding{53} &  $\star\star\star$ & Official (C++)\footnote{\url{https://github.com/rupea/LabelFilters}}\\
			\cline{2-2} \cline{8-8}         & PD-Sparse~\cite{yen2016pd} 			&     \cellcolor[gray]{0.9} $\triangle$  &   \cellcolor{pink} \ding{53}    &     \cellcolor{lavenderblue} $\bigcirc$  &       \cellcolor{pink} \ding{53}  &  $\star\star$ & Official (C++)\footnote{\url{https://github.com/a061105/ExtremeMulticlass}}\\
			\cline{2-2} \cline{8-8}         & PPD-Sparse~\cite{yen2017ppd} 							&      \cellcolor{lavenderblue} $\bigcirc$ &    \cellcolor{pink} \ding{53}   &      \cellcolor[gray]{0.9} $\triangle$ &      \cellcolor{pink} \ding{53} &  $\star\star\star$ & Official (C++)\footnote{\url{http://manikvarma.org/code/Parabel/download.html}}\\
			\cline{2-2} \cline{8-8}         & Parabel~\cite{prabhu2018parabel} 						&    \cellcolor{lavenderblue} $\bigcirc$   &   \cellcolor{pink} \ding{53}    &   \cellcolor{lavenderblue} $\bigcirc$    &       \cellcolor{pink} \ding{53}     &  $\star\star\star\star$ & Official (C++)\footnote{\url{http://manikvarma.org/code/Parabel/download.html}}\\
			\cline{2-2} \cline{8-8}         & ProXML~\cite{babbar2019data} 						&      \cellcolor[gray]{0.9} $\triangle$ &    \cellcolor{lavenderblue} $\bigcirc$   &      \cellcolor[gray]{0.9} $\triangle$ &           \cellcolor{pink} \ding{53} &  $\star\star\star\star$ & Official (C++)\footnote{\url{https://github.com/xmc-aalto/proxml}}\\
			\cline{2-2} \cline{8-8}         &
			Bonsai~\cite{khandagale2019bonsai} &  \cellcolor{lavenderblue} $\bigcirc$ & \cellcolor{pink} \ding{53} & \cellcolor{lavenderblue} $\bigcirc$ & \cellcolor{pink} \ding{53} &  $\star\star\star\star$ & Official (C++)\footnote{\url{https://github.com/xmc-aalto/bonsai}}\\
			\cline{2-2}  \cline{8-8}        & Slice~\cite{DBLP:conf/wsdm/JainBCV19} 				&     \cellcolor{lavenderblue} $\bigcirc$  &   \cellcolor{pink} \ding{53}    &         \cellcolor{lavenderblue} $\bigcirc$ &      \cellcolor{pink} \ding{53} &  $\star\star\star$ & Official (C++)\footnote{\url{http://manikvarma.org/code/Slice/download.html}}\\
			\midrule
			\multirow{16}{*}{Deep Learning} 
			& MACH~\cite{medini2019extreme}			&       \cellcolor{lavenderblue} $\bigcirc$&       \cellcolor{pink} \ding{53}&       \cellcolor{pink} \ding{53}&         \cellcolor{pink} \ding{53}& $\star\star$ & Official(Python)\footnote{\url{https://github.com/Tharun24/MACH}}  \\
			\cline{2-2}  \cline{8-8}
			& GLaS~\cite{chuanguo2019nips} 							&    \cellcolor{pink} \ding{53}    &   \cellcolor{lavenderblue} $\bigcirc$    &       \cellcolor[gray]{0.9} $\triangle$       &   \cellcolor{pink} \ding{53}    &  $\star\star\star\star$ & Unofficial (Python)\footnote{\url{ https://github.com/Stomach-ache/GLaS}} \\
			\cline{2-2}  \cline{8-8}
			& fastTextLearnTree~\cite{Jernite2017fasttextlearntree} 			&      \cellcolor{lavenderblue} $\bigcirc$ &   \cellcolor{pink} \ding{53}    &     \cellcolor{lavenderblue} $\bigcirc$  &           \cellcolor{pink} \ding{53} & $\star\star$ & Official (Python)\footnote{\url{https://github.com/yjernite/fastTextLearnTree}}\\
			\cline{2-2}  \cline{8-8}
			& XML-CNN~\cite{DBLP:conf/sigir/LiuCWY17} 				&     \cellcolor{pink} \ding{53} &   \cellcolor{pink} \ding{53}    &       \cellcolor{lavenderblue} $\bigcirc$ &          \cellcolor{pink} \ding{53} & $\star$ & Official (Python)\footnote{\url{https://github.com/siddsax/XML-CNN}}\\
			\cline{2-2}  \cline{8-8}
			& AttentionXML~\cite{you2018attentionxml} 				&       \cellcolor{pink} \ding{53}&	\cellcolor{lavenderblue} $\bigcirc$&       \cellcolor{pink} \ding{53}&          \cellcolor{pink} \ding{53}&       $\star\star\star\star$ &Official (Python)\footnote{\url{https://github.com/yourh/AttentionXML}} \\
			\cline{2-2}  \cline{8-8}
			& LightXML~\cite{you2018attentionxml} 				&       \cellcolor{pink} \ding{53}&	\cellcolor{lavenderblue} $\bigcirc$&       \cellcolor{pink} \ding{53}&          \cellcolor{pink} \ding{53}&       $\star\star\star\star$ &Official (Python)\footnote{\url{http://github.com/kongds/LightXML}} \\
			\cline{2-2}  \cline{8-8}
			
			& APLC\_XLNet~\cite{DBLP:journals/corr/abs-2007-02439}  &    \cellcolor[gray]{0.9} $\triangle$  &    \cellcolor{lavenderblue} $\bigcirc$   &      \cellcolor{lavenderblue} $\bigcirc$ &       \cellcolor{pink} \ding{53}  & $\star\star\star\star$ & Official (Pyton)\footnote{\url{https://github.com/huiyegit/APLC_XLNet}} \\
			\cline{2-2}  \cline{8-8}
			& X-Transformer~\cite{chang2019x} 						&     \cellcolor{pink} \ding{53} &   \cellcolor{pink} \ding{53}    &      \cellcolor{pink} \ding{53} &          \cellcolor{pink} \ding{53} & $\star\star\star$ & Official (Python)\footnote{\url{https://github.com/OctoberChang/X-Transformer}} \\
			\cline{2-2}  \cline{8-8}
			& XR-Linear~\cite{yu2022pecos} 							&       \cellcolor{lavenderblue} $\bigcirc$ &       \cellcolor{lavenderblue} $\bigcirc$ &       \cellcolor[gray]{0.9} $\triangle$ &         \cellcolor{pink} \ding{53}  & $\star\star\star\star$  & Official(C++,Python)\footnote{\url{https://github.com/amzn/pecos}}\\
			\cline{2-2}  \cline{8-8}
			& XR-Transformer~\cite{zhang2021fast} 					&       \cellcolor{pink} \ding{53} &    \cellcolor{lavenderblue} $\bigcirc$ &      \cellcolor{pink} \ding{53} &           \cellcolor{pink} \ding{53} & $\star\star\star\star$ & Official(C++,Python)\footnote{\url{https://github.com/amzn/pecos}} \\
			\cline{2-2}  \cline{8-8}
			& ZestXML~\cite{gupta2021generalized} 					&       \cellcolor{lavenderblue} $\bigcirc$ &  \cellcolor[gray]{0.9} $\triangle$ & 		       \cellcolor{lavenderblue} $\bigcirc$ &        \cellcolor{lavenderblue} $\bigcirc$ & $\star\star\star\star$& Official(C++,Python)\footnote{\url{https://github.com/nilesh2797/zestxml}} \\
			\cline{2-2}  \cline{8-8}
			& MACLR~\cite{xiong2021extreme} 						&       	\cellcolor{lavenderblue} $\bigcirc$&   \cellcolor{pink} \ding{53} & 	         \cellcolor[gray]{0.9} $\triangle$ &       \cellcolor{lavenderblue} $\bigcirc$ & $\star\star\star\star$ &  Official(Python)\footnote{\url{https://github.com/amzn/pecos/tree/mainline/examples/MACLR}} \\
			\cline{2-2}  \cline{8-8}
			& DECAF~\cite{mittal2021decaf} 							&\cellcolor[gray]{0.9} $\triangle$&       \cellcolor{lavenderblue} $\bigcirc$&       \cellcolor[gray]{0.9} $\triangle$&       \cellcolor{pink} \ding{53}& $\star\star\star\star$&Official(Python)\footnote{\url{https://github.com/Extreme-classification/DECAF}} \\
			\cline{2-2}  \cline{8-8}
			& SiameseXML~\cite{dahiya2021siamesexml} 				&       \cellcolor[gray]{0.9} $\triangle$ &    \cellcolor{lavenderblue} $\bigcirc$ &       \cellcolor{pink} \ding{53}&           \cellcolor{pink} \ding{53}& $\star\star\star\star\star$&Official(Python)\footnote{\url{https://github.com/Extreme-classification/siamesexml}} \\
			\cline{2-2}  \cline{8-8}
			& Astec~\cite{dahiya2021deepxml} 						&       \cellcolor[gray]{0.9} $\triangle$&    \cellcolor{lavenderblue} $\bigcirc$&       \cellcolor{pink} \ding{53}&         \cellcolor{pink} \ding{53}& $\star\star\star\star$&Official(Python)\footnote{\url{https://github.com/Extreme-classification/deepxml}}  \\
			\cline{2-2}  \cline{8-8}
			& ECLARE~\cite{mittal2021eclare} 						&        \cellcolor[gray]{0.9} $\triangle$&    \cellcolor{lavenderblue} $\bigcirc$   &        \cellcolor[gray]{0.9} $\triangle$&           \cellcolor{pink} \ding{53} & $\star\star\star\star\star$ & Official(Python)\footnote{\url{https://github.com/Extreme-classification/ECLARE}}\\
			\cline{2-2}  \cline{8-8}
			& GalaXC~\cite{saini2021galaxc} 						&       \cellcolor[gray]{0.9} $\triangle$&   \cellcolor{lavenderblue} $\bigcirc$&       \cellcolor[gray]{0.9} $\triangle$&             \cellcolor{pink} \ding{53}& $\star\star\star\star\star$  & Official(Python)\footnote{\url{https://github.com/Extreme-classification/GalaXC}}\\
			\bottomrule
			\bottomrule
		\end{tabular}
\end{table*}%
\end{savenotes}

\begin{figure*}
\centering
\includegraphics[width=1\textwidth]{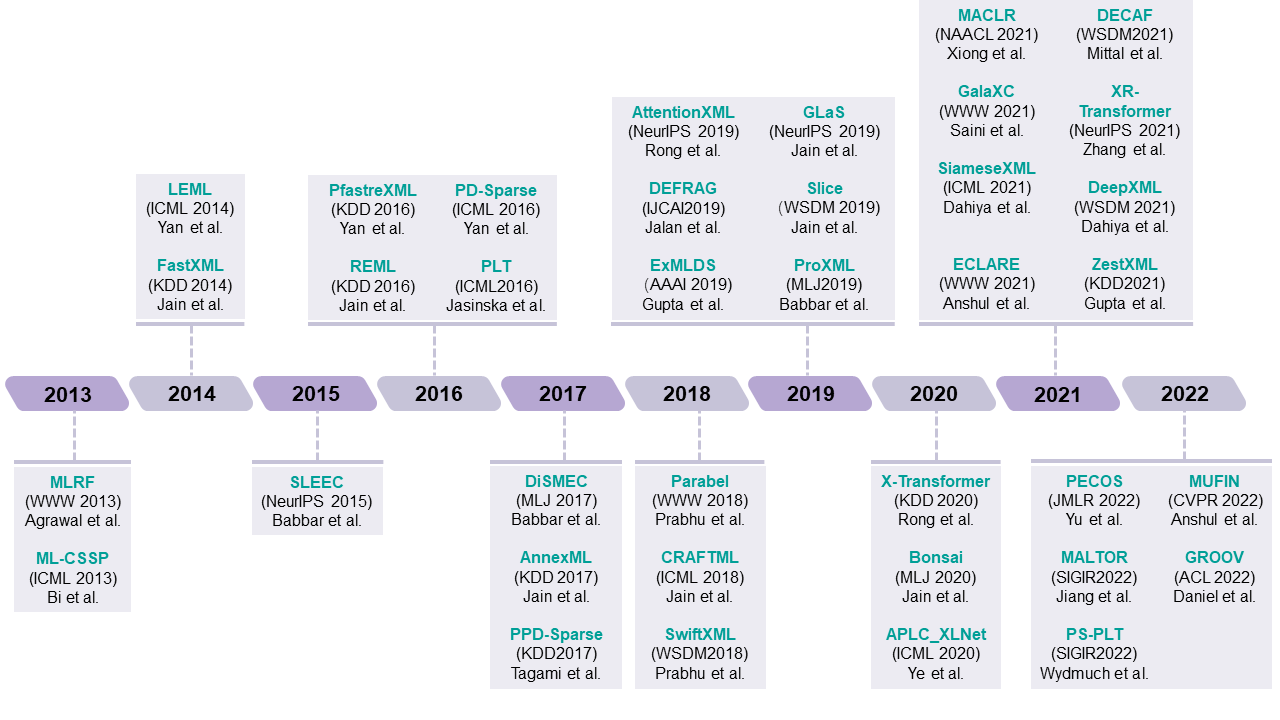}
\caption{A chronological overview of recent representative work in XML.}
\label{fig:timeline}
\end{figure*}

\subsection{Commonly Used Benchmark Datasets}
\begin{itemize}
\item  \textbf{Bibtex}~\cite{katakis2008multilabel}: It is collected from a social bookmarking and publication-sharing system named Bibsonomy, where a user may store and organize Bookmarks (web pages) and BibTeX entries. In this dataset, each instance represents a BibTeX item submitted to Bibsonomy. Metadata for the BibTeX item, like the title of the paper, the authors, and so on, are extracted as features. The tags assigned by users to the item is used as labels.


\item \textbf{Delicious}~\cite{tsoumakas2008effective}: It is collected from the social bookmarking site Delicious. Each instance represents a webpage bookmark, from which high-frequency words are extracted as features, and the popular tags listed in the web page are used as labels.

\item \textbf{Mediamill}~\cite{DBLP:conf/mm/SnoekWGGS06}: Each instance represents a camera shot from broadcast news, and the task is to learn semantic concept contained in this shot. The input feature vector is yielded from a key frame image, representing scores of similarity between some predefined regions in the key frame image and some low-level visual concepts, like road, sky, water body, and so on.

\item \textbf{EUR-Lex}~\cite{mencia2008efficient,DBLP:journals/jmlr/TsoumakasXVV11,DBLP:conf/acl/ChalkidisFMA19,DBLP:journals/corr/abs-1905-10892}: It is collected from many different types of documents about European Union Law. For each document, the tokens extracted from the text body, together with their term frequency, are used as features, and the descriptors in bibliographic notes of the document are extracted as labels.

\item \textbf{Amazon}~\cite{DBLP:conf/recsys/McAuleyL13,DBLP:conf/sigir/McAuleyTSH15}:
each instance represents an item (usually a book) identified with a unique item id. The raw instance feature are website contents (usually, the instance feature we use in experiments except in deep learning are processed using natural language processing techniques, such as one-hot encoding), assume the items are books, including book id, title, author, consumer reviews and other information.

\item \textbf{Wikipedia}~\cite{zubiaga2012enhancing,DBLP:journals/corr/PartalasKBAPGAA15}:
The raw instance feature are website contents and the meta-label of this web page is the categories that this page belongs to.
\end{itemize}

\textbf{Text Feature}
In XML, the feature space is usually very sparse and high dimensional, such as the commonly used bag-of-words features in text classification. To alleviate this issue, an alternative is using dense embedding obtained by NLP tools, such as GloVe~\cite{pennington2014glove} and Transformer~\cite{chang2019x}. Recently X-bert~\cite{chang2019x} proposes to apply BERT~\cite{DBLP:conf/naacl/DevlinCLT19bert,liu2019roberta} to XML, which could better capture semantic meaning of documents. To make BERT feasible in large-scale problems, they first group labels into clusters leveraging label semantic information. Documents are then mapped to clusters of relevant through a multi-label classifier. Finally, binary relevance~\cite{rifkin2004onevsall} classifiers are employed within each cluster to rank individual labels. Similarly, APLC-XLNet~\cite{DBLP:journals/corr/abs-2007-02439} uses XLNet~\cite{yang2019xlnet} to model context information instead of BERT. AttentionXML~\cite{you2018attentionxml} uses GloVe embedding as the input of BiLSTM, which achieves the current state-of-the-art results.

\subsection{Evaluation Metrics}
In this section, we presents the conventional evaluation metrics used to assess XML models with a test dataset.
\begin{itemize}
\item  \textbf{P@$k$} is a commonly used ranking based performance measure in XML and has been widely adopted for ranking tasks~\cite{Prabhu2014fastxml,NIPS2015sleec}. In Top-$k$ precision, only a few top predictions of an instance will be considered. For each instance $\x_i$, the Top-$k$ precision is defined for a predicted score vector $ \hat{\y}_i \in \mathbb{R}^{L}$ and ground truth label vector $\y_i \in \{-1, 1\}^{L}$ as
\begin{equation}
	\text{P@}k := \frac{1}{k} \sum_{l\in \text{rank}_k (\hat{\boldsymbol y})} \boldsymbol y_l,
\end{equation}
where $\mathrm{rank}_{k}(\hat{\y}_i)$ returns the indices of $k$ largest value in $\hat{\y}_i$ ranked in descending order. 

\item \textbf{nDCG@$k$} is another commonly used ranking based performance measure and is defined as
\begin{equation}
	\text{nDCG@}k := \frac{\text{DCG@}k}{\sum_{l=1}^{\min(k, \|\boldsymbol y\|_0)} \frac{1}{\log(l+1)}},
\end{equation}
where $\text{DCG@}k := \sum_{l\in {\text{rank}}_k (\hat{\boldsymbol y})} \frac{\boldsymbol y_l}{\log(l+1)}$ and $||\boldsymbol{y}||_{0}$ returns the 0-norm of the true-label vector.

\item \textbf{PSP@$k$}
Propensity scored variants of such losses, including precision@k and nDCG@k, are developed and proved to give unbiased estimates of the true loss function even when ground
truth labels go missing under arbitrary probabilistic label
noise models~\cite{jain2016propensity}.
\begin{equation}
	\text{PSP}@k := \frac{1}{k} \sum_{l\in \text{rank}_k (\hat{\boldsymbol y})} \frac{\boldsymbol y_l}{p_l}
\end{equation}
$p_{l}$ is the propensity score for label $l$ which helps in making metrics unbiased.

\item \textbf{PSnDCG@$k$}
\begin{equation}
	\text{PSnDCG}@k := \frac{{\text{PSDCG}}@k}{\sum_{l=1}^{k} \frac{1}{\log(l+1) } }
\end{equation}
where
$\text{PSDCG}@k := \sum_{l\in {\text{rank}}_k (\hat{\boldsymbol y})} \frac{\boldsymbol y_l}{p_l\log(l+1)}$. Note that $p_l$ represents the inverse propensity score for the $l$-th label, which is defined as:
$$
p_{l} = 1+C\left(n_{l}+B\right)^{-A},
$$
and $A, B, C$ are hyperparameters which can be set as recommended in paper~\cite{jain2016propensity}.
\end{itemize}

\underline{\textbf{Remark:}}
Although the propensity score based performance metrics (PSP@$k$ and PSnDCG@$k$) proposed in PfastreXML~\cite{jain2016propensity} can, to a certain extent, alleviate this problem, the parameters $A$, $B$, and $C$ are set in a heuristic manner when calculating label propensities for the given data set.
Opposed to PfastreXML~\cite{jain2016propensity}, ProXML~\cite{babbar2018adversarial} claimed that hamming loss is a more suitable loss function to optimize since it takes all labels into account during optimization instead of only the top $k$ predictions considered in PSP@$k$ and PSnDCG@$k$.
Additionally, to increase the diversity of predicted labels, we adapt the term $coverage$,
which has been widely used in recommender systems to measure the diversity of recommended items~\cite{ge2010beyond,castells2015novelty}.
Coverage is defined as the fraction of items that appear in the users' recommendation lists.
In the context of XML, we refer to the $coverage$ as:
\begin{equation*}
Coverage = \frac{|\cup_{1\leq i \leq N}\mathcal{R}_i|}{L}
\end{equation*}
where $\mathcal{R}_i$ is the set of relevant labels of instance $i$ correctly annotated.
However, directly optimizing the $coverage$ objective is very difficult, and we leave this problem for future work. Besides, MALTOR~\cite{DBLP:conf/sigir/JiangCZHY22} considers that fixed length estimation ignores other relevant labels and learns best length from documents to improve prediction accuracy.

\section{Future Research Directions}\label{sec-future-work}
With the rapid growth of data scale in real-world applications, XML faces several challenges. This section discusses the limitations of the existing XML methods and provides potential directions for future research that can facilitate and envision the development of XML.

\subsection{Problems of Existing Methodologies}
\begin{enumerate}
\item \textit{Tree-based Method.} An important issue of tree-based method is that it does not optimize the global performance in each node partitioning. Many approaches apply clustering to split label set into several subsets for its simplicity and efficiency. However, it still needs further study if such clustering is optimal.
\item \textit{Embedding-based Method.} GLaS~\cite{chuanguo2019nips} argues that deep embedding-based methods usually suffer from over-fitting, which is the root cause of their inferior performance compared with other types of approaches. However, it is also observed that methods as simple as linear classifiers~\cite{babbar2017dismec} can over-fit. In addition, it is also not verified if traditional embedding-based methods, such as LEML~\cite{yu2014leml} and SLEEC~\cite{NIPS2015sleec}, over-fit the data.
\end{enumerate}

\subsection{The Tail Label Problem}
\begin{enumerate}
\item \textit{How to split the label set into head and tail?} It is known that labels follows a long-tailed distribution on XML datasets. Some prior work~\cite{dahiya2021deepxml} attempts to transfer knowledge from data-rich head to data-scarce tail labels, which needs to partition the label set into head and tail in ahead. However, there is no principle way of partitioning the entire label set into head and tail labels. In particular, a threshold is needed for the split. Existing methods either set a particular fraction of labels as tail~\cite{wei2018ijcai} or according to the label frequency~\cite{DBLP:conf/iclr/KangXRYGFK20}. By partitioning the label set, one can model head and tail labels separately using different methodologies. Thus, it is a fundamental problem to study the split of head and tail labels.
\item  \textit{How to do trade-off between head and tail?} As shown in previous work~\cite{jain2016propensity}, the metric P@$k$ and PSP@$k$ cannot be simultaneous optimized. Moreover, when the PSP@$k$ is optimized, the P@$k$ drops. The reason is that P@$k$ hinges on more accurate predictions on head labels, while optimizing P@$k$ might hurt the accuracy on head labels. To this end, a trade-off policy is desired according to specific XML tasks.
\item \textit{How to design loss functions for tail labels?} Although re-weighted losses are the most popular ways to optimize in the literature~\cite{jain2016propensity,lin2017focal,DBLP:conf/iclr/KangXRYGFK20}, recent work~\cite{babbar2018adversarial} points out that it can achieve comparable results by optimizing hamming loss, which is deemed as unsuitable for multi-label learning. As such, appropriately re-weighting  tail labels is still an open problem in future studies.
\item \textit{How to evaluate models in presence of tail labels?} Conventionally, evaluation metrics are calculated by directly matching up the predicted labels $\hat{\y}$ and ground-truth labels $\y$. Even for propensity scored metrics, e.g., PSP@k and PSnDCG@k, it cannot explicitly see the contribution of tail labels to the performance. Schultheis et al.~\cite{DBLP:journals/corr/abs-2207-13186} pointed out that The current propensity metrics seem to evaluate both tail labels and missing labels together. Besides, current propensity model~\cite{jain2016propensity} has some shortcomings which lead to deviation from ground truth score. Instead, the label set could be split into multiple partitions, e.g., \textit{head}, \textit{body}, and \textit{tail}, according to label frequencies. Given predicted labels $\hat{\y}$, its performance on each of the label partitions should be measured. By doing this, it is clear to see how does the model perform on head labels and tail labels.
\end{enumerate}

\subsection{Weakly Supervised XML}
\begin{enumerate}
\item \textit{Extreme multi-instance multi-label learning.} Recently, EAGLE~\cite{DBLP:journals/corr/abs-2004-00198} studies a new XML setting for the first time, i.e., extreme multi-instance multi-label learning. This problem has many real-world applications. For instance, in video classification, tags are labeled in video levels, one may want to predict relevant tags for each frame of the video. 
The design of EAGLE follows the simple principle that the label embedding should be close to the embedding of at least one of the sample points in every positively labeled group. EAGLE can be seen a baseline for this problem while leaves big room for performance improvement.

\item \textit{Open-set extreme multi-label learning.} In many real-world applications, new data is generated very fast and emerging labels out of the observed label set might be seen, i.e., open-set labels. Thus, one needs to learn to detect open-set labels in addition to classify known labels. 

\item \textit{Extreme multi-label learning with streaming labels.}
Multi-label learning with streaming labels has been studied in several previous works~\cite{DBLP:journals/corr/YouXW0T16,zhu2018multi,icml2020_dsll}. To deal with streaming labels, one usually needs to achieve online updates for trained models. However, existing approaches either cannot scale to XML problems or assume new labels are known in prior which may not be a good approximation of the real-world streaming labels. Thus, 

\item \textit{Extreme multi-label learning with missing labels.}
Owing to the large label space, it is costly to perfectly label each sample with its relevant labels. Therefore, extreme multi-label learning in the presence of missing labels is prevalent. One way to handle missing labels is to develop unbiased loss functions as done in some previous studies. It would be also interesting to mitigate the influence of missing labels via label enhancement, e.g., label distribution learning~\cite{DBLP:journals/tkde/Geng16}. As a special case of missing labels, zero-shot XML has drawn a lot of attention, in which it is more challenging to distinguish tail labels and unseen labels.



\end{enumerate}

\subsection{Modeling Label Correlations}
\begin{enumerate}
\item \textit{How to explicitly model label correlations?} Exploiting the correlation among labels is the most important problem in multi-label learning, which differs multi-label learning from multi-class learning. Existing works explore label correlation implicitly by enforcing pairwise ranking loss~\cite{chuanguo2019nips} and label clustering~\cite{khandagale2019bonsai}. On the opposite, some other works attempt to explicitly learn label correlations~\cite{zhang2010labeld,DBLP:conf/kdd/Papagiannopoulou15} using Bayesian network, while it cannot naturally scale to XML problems due to the huge computation overhead. Thus, it is interesting to develop an efficient way of exploiting label corrections can be helpful in XML. 
\item \textit{How to incorporate domain knowledge?} Due to the large label space and the scarcity of data, it may be very challenging to directly learn correlations among labels, therefore one walk around is to incorporate domain knowledge~\cite{fischer2019dl2}. Then, the provided knowledge can act as constraints to optimizing objectives or the modeling of label corrections. For instance, it is a common practice to borrow knowledge from Wikipedia to calculate label similarities. Thus, it still has big room for improvement in XML via leveraging external knowledge.
\end{enumerate}

\subsection{XML Meets Computer Vision}
Almost existing studies in XML use text data for model evaluation, while image data is another common modality which has not been explored. Moreover, long-tailed image recognition has been studied in many recent computer vision literature~\cite{deng2010does,DBLP:conf/nips/CaoWGAM19,DBLP:conf/iclr/KangXRYGFK20,DBLP:conf/nips/WangRH17,DBLP:journals/corr/abs-1912-02413,DBLP:journals/corr/abs-2001-01536}, and several long-tailed image classification benchmark datasets are crafted~\cite{iNatrualist,DBLP:conf/cvpr/0002MZWGY19}. As one of the most frequently used techniques, class-balanced sampling is applied to mitigate the negative influence of class imbalance which can improve the generalization of head classes. However, implementing class-balanced sampling is not straightforward in multi-labeled data. MLUL~\cite{DBLP:journals/pr/LiuBT22} recently makes a step towards this direction and achieve superior performance than existing works by eliminating local imbalance using multi-label sampling approaches. Other techniques, e.g., logit adjustment~\cite{DBLP:journals/corr/abs-2007-07314}, label-distribution-aware margin loss~\cite{DBLP:conf/nips/CaoWGAM19}, and decoupled training~\cite{DBLP:conf/iclr/KangXRYGFK20}, have not been examined in XML yet, and it would be interesting to adapt multi-class image recognition methods to multi-label text classification. Additionally, it is also a realistic problem tTo simultaneously handle both text and image data, MUFIN~\cite{mittal2022multi} makeis the first attempt in this problemto apply XML ideas to multi-modal problems, broadening the boundaries of XML.


\section{Conclusion}\label{sec-conclude}
Conventional multi-label learning usually deals with small-scale datasts, while the data scale rapidly increases on the internet. Even worse, in real-world applications, the frequency of labels often follows a long-tailed distribution, making the problem more challenging. Extreme multi-label learning was proposed to deal with these difficulties.

In this survey, we present a comprehensive overview of XML, including background knowledge, main challenges, model architectures, various extensions, publicly available resources, and evaluation metrics. Furthermore, a systematic comparison was conducted using five popular properties used for evaluation in the recent literature. According to the comparison results, there is no ideal method that supports all the required properties; the supported properties varied depending on the category to which each method belonged. Finally, we suggest several possible future research directions for XML.





\ifCLASSOPTIONcaptionsoff
\newpage
\fi



\bibliographystyle{IEEEtran}
\bibliography{IEEEabrv,citation}

%




\end{document}